\documentclass[journal]{IEEEtran}
\usepackage{amsmath,epsfig,color}
\usepackage{url}
\usepackage{bigstrut,multirow,rotating}
\usepackage{textcomp,booktabs}
\usepackage{tabularx}
\usepackage{floatrow}
\newfloatcommand{capbtabbox}{table}[][\FBwidth]
\usepackage{colortbl}
\usepackage{xcolor}
\usepackage{multirow}
\usepackage{booktabs}
\usepackage{underscore}
\usepackage[marginal]{footmisc}
\usepackage{subfigure}
\usepackage{algorithm,algorithmic}
\usepackage{amssymb}
\begin{document}
\newcommand{\bl}[1]{{\color{black}#1}}
\newcommand{\bbl}[1]{{\color{black}#1}}
\title{No-reference Screen Content Image Quality Assessment with Unsupervised Domain Adaptation}
\author{Baoliang Chen, Haoliang Li,  Hongfei Fan and Shiqi Wang,~\IEEEmembership{Member,~IEEE}
\thanks{B. Chen and S. Wang are with the Department of Computer Science, City University of Hong Kong, Hong Kong (e-mail: blchen6-c@my.cityu.edu.hk; shiqwang@cityu.edu.hk). Corresponding author: Shiqi Wang. }
\thanks{H. Li is with the Department of Electrical Engineering, City University of Hong Kong (email: haoliang.li1991@gmail.com).}
\thanks{H. Fan is with Kingsoft Cloud, China (email: fanhongfei@kingsoft.com).}
}

\maketitle

\begin{abstract}
In this paper, we quest the capability of transferring the quality of natural scene images to the images that are not acquired by optical cameras (e.g., screen content images, SCIs), rooted in the widely accepted view that the human visual system has adapted and evolved through the perception of natural environment. Here, we develop the first unsupervised domain adaptation based no reference quality assessment method for SCIs, leveraging rich subjective ratings of the natural images (NIs). In general, it is a non-trivial task to directly transfer the quality prediction model from NIs to a new type of content (i.e., SCIs) that holds dramatically different statistical characteristics. Inspired by the transferability of pair-wise relationship, the proposed quality measure operates based on the philosophy of 
improving the transferability and discriminability simultaneously. In particular,
we introduce three types of losses which  complementarily and explicitly regularize the feature space of ranking in a progressive manner.
Regarding feature discriminatory capability enhancement, we propose a center based loss to rectify the classifier and improve its prediction capability not only for source domain (NI) but also the target domain (SCI). For feature discrepancy minimization, the maximum mean discrepancy (MMD) is imposed on the extracted ranking features of NIs and SCIs. Furthermore, to further enhance the feature diversity, we introduce the correlation  penalization between different feature dimensions, leading to the features with lower rank and higher diversity.
Experiments show that our method can achieve higher performance on different source-target settings based on a light-weight convolution neural network.
The proposed method also sheds light on learning quality assessment measures for unseen application-specific content without the cumbersome and costing subjective evaluations.

\end{abstract}

\IEEEpeerreviewmaketitle

\section{Introduction}
\IEEEPARstart{R}{ecent} years have witnessed the surge of applications based on deep learning, which primarily rely on training of neural networks with large-scale labelled data. No reference image quality assessment (NRIQA), which operates without the pristine reference, has also greatly benefited from such a principled pipeline~\cite{moorthy2011blind,gu2014using,mittal2012no,kang2014convolutional}. \bl{Generally speaking, early NRIQA methods are mainly based on the design of hand-crafted features~\cite{moorthy2011blind,liu2018blind,wu2015blind}.  The free-energy principle proposed in brain theory and neuroscience~\cite{friston2006free,friston2010free,zhai2011psychovisual,zhai2019free} reveals that the image distortion can be measured by the discrepancy between the image and its brain predicted version. Natural scene statistics (NSS) features extraction~\cite{liu2019unsupervised,mittal2012no, zhai2019free} has also become a feasible way  and the quality prediction can be learned in an unsupervised manner.
As deep learning technologies achieved great success in various computer vision tasks, convolutional neural network (CNN)-based features are widely introduced in BIQA~\cite{kang2014convolutional,min2020study,kim2016fully,ma2017end,zhu2020multiple,zhang2021uncertainty,lu2020automatic} which can achieve higher prediction performance on different IQA databases.} However, the strong assumption that the training and testing data are drawn from closely aligned feature spaces and distributions creates the risk of poor generalization capability, and as a consequence, inaccurate predictions are obtained on the images that hold dramatically different statistics compared to those in the training set. However, in real applications scenarios, obtaining the ground-truth label via subjective testing 
for any content of interest is quite time-consuming and cumbersome. As such, though there are numerous labelled image quality assessment datasets based on natural images, the emergence of image creation routines and methods brings new challenges on the NRIQA based on the available labelled data.

In this paper, we answer the question that whether the quality of natural scene could be transferable by designing a novel NRIQA method based on unsupervised domain adaptation (UDA). There are three reasons behind which the natural images are treated as the source domain in this study. First, there are numerous datasets of natural images with subjective ratings, and the need for transferring from statistically regular natural images to other types of content is becoming more pronounced. Second, there is an increasing consensus that the human visual system (HVS) evolves with the natural scene statistics, such that transferring from natural to unnatural scene could closely resemble human behavioral responses when evaluating the quality of artificially created images. Third, it is widely acknowledged that the  artificially created images do not follow the natural scene statistics, such that it is meaningful to investigate the transfer capability 
and this methodology could be feasibly extended to many other scenarios. As such, in contrast to natural images (NIs) which form the source domain, we choose screen  content imags (SCIs) as the target domain. It has been repeatedly validated that SCIs hold dramatically different statistics compared to NIs~\cite{wang2016just,min2017unified}.
Therefore, given the subjective ratings of the natural images only, it is a quite challenging task to transfer the quality of natural scenes to the unnatural screen content.   


The transferability of quality prediction differs substantially from other computer vision tasks (e.g., object and action recognition). Quality assessment, the aim of which is matching human measurements of perceptual quality, highly relies on the image content. To tackle this problem, we propose to leverage the advantage of domain adaptation (DA), in an effort to learn a NRIQA model specifically for SCIs (target domain) from the NIs (source domain)
as well as corresponding ground-truth ratings of NIs. This scenario falls into the unsupervised domain adaptation which has been widely studied in the literature~\cite{ben2007analysis,ben2010theory,fernando2013unsupervised}. However, the difficulty of directly transferring the quality prediction model from NIs to SCIs arises due to underlying differences in terms of their characteristics. Instead of 
straightforwardly training the model with the labelled data to equip the capability of quality prediction, we propose to explore the transferability of pair-wise relationship by learning to rank, 
such that the model that infers the quality rank of a pair of images can be learnt. More concretely, discriminable image pairs from source and target domains are selected to learn the ranking model. Grounded on the work of embedding DA in the the process of representation learning, a feature that accounts for the ranking equipping with the property of domain invariant is expected to be learnt, such that with the reduction of domain shift, the knowledge learned in the source domain can be well transferred to the target domain, leading to signiﬁcant performance improvement in target domain quality assessment. To this end, we introduce three complementary losses to explicitly regularize the feature space of pair-wise relationship in a progressive manner. The main contributions of this paper are as follows,
\begin{itemize}
\item We propose a novel framework to learn the discriminative and transferable pair-wise quality relationship, in an effort to tackle the problem of NRIQA for SCIs with UDA. The pairwise ranking finally leads to the predicted quality with the quality regression network, which has been validated to achieve higher prediction accuracy compared to the state-of-the-art models. 
\item For the source domain, we fully leverage the label information at image and pair levels to learn the features with strong discriminability. Moreover,  we encourage our model to learn the quality feature with high diversity by penalizing the correlation between different dimensions, leading to features with reduced redundancy and higher generalization capability. 

\item For the target domain, to reduce the feature discrepancy between the two domains and improve the transferability capability, the maximum mean discrepancy (MMD) is imposed for aligning the two domains. For  better  prediction  performance  on  the  target  domain,  we further rectify the classifier  by  a  center  based  rectification  module. 

\end{itemize}

\section{Related Works}
\subsection{No-Reference Image Quality Assessment} 

Conventional NRIQA methods rely on the assumption that the natural scene statistics (NSS) are governing the perception and behavior of natural images, such that the distortion is reflected by the destruction from naturalness. In \cite{moorthy2011blind}, based on NSS in wavelet domain, the un-naturalness of distorted images is characterized. Saad \textit{et al.} established the NSS model in discrete cosine transform (DCT) domain and the quality is predicted by the Bayesian inference~\cite{saad2012blind}. Different from quality regression, Hou \textit{et al.} designed a deep learning model to classify the NSS features into five grades, corresponding 
to five quality levels \cite{hou2014blind}. Generally speaking, the deep learning based methods rely on large-scale training samples with subjective ratings as the label information \cite{kang2014convolutional,kim2016fully,bosse2017deep,bianco2018use,gu2019blind,fu2016blind,kim2018multiple}. However, due to the insufficient  training data, extra synthetic databases have also been taken advantage of \cite{zhang2018blind,ma2017end}, in which the distortion type identification network can be identified as ``prior knowledge'' and combined with the quality prediction network. Different from learning with a single image, ranking based methods \cite{liu2017rankiqa} 
have also been proposed to enrich the training data. In general, to straightforwardly acquire the rank information, the image content within one pair is usually required to be identical, lacking the capacity of cross content quality prediction. 
 
Due to the distinct statistics of SCIs, numerous NRIQA methods have been specifically developed. In \cite{gu2017no}, four types of features including picture complexity, screen content statistics, global brightness and sharpness of details are extracted for SCI quality prediction. In \cite{fang2017no}, inspired by the perception characteristics, Fang \textit{et al.} designed a quality assessment method by 
the combination of local and global texture features and luminance features. Driven by the hypothesis that HVS is highly sensitive to sharp edges, in \cite{zheng2019no}, the regions in SCI are divided into sharp edge regions and non-sharp edge regions, such that the hybrid region based features are extracted for no-reference SCI quality assessment. Benefiting from the powerful feature extraction capability of CNNs, Zuo \textit{et al.} proposed a CNN based framework with two sub-networks, where one of the sub-network is designed for producing the
local quality of each image patch and the other is responsible for image level quality fusion \cite{zuo2016screen}.

\bl{In the literature, developing a unified IQA model that is applicable across different content types have also attracted increasing attentions. In~\cite{min2018saliency}, a saliency-induced reduced-reference (SIRR) IQA measure was proposed for the quality measure of both NIs and SCIs, where the quality can be estimated by comparing the similarity between two  saliency maps.  Min \textit{et al.} \cite{min2017unified} proposed an unified content-type adaptive (UCA) BIQA model that is applicable across content types by integrating a content-adaptive multi-scale weighting strategy. In 
contrary to the conventional IQA measures, in \cite{min2017blind}, a pseudo-reference image
(PRI) based BIQA model was proposed and the quality can be estimated by measuring the distance to the PRI. In particular, several PRI-based distortion-specific measures are  constructed in this model to estimate blockiness, sharpness and noisiness.  This idea was further developed in~\cite{min2018blind}, where multiple pseudo reference images (MPRIs) are generated by degrading the distorted image. As such, the influence of image content can be mitigated, leading to more accurate and consistent inference of the image quality.}

\begin{figure*}[t]
\begin{minipage}[b]{0.95\linewidth}
  \centering
  \centerline{\includegraphics[width=1\linewidth]{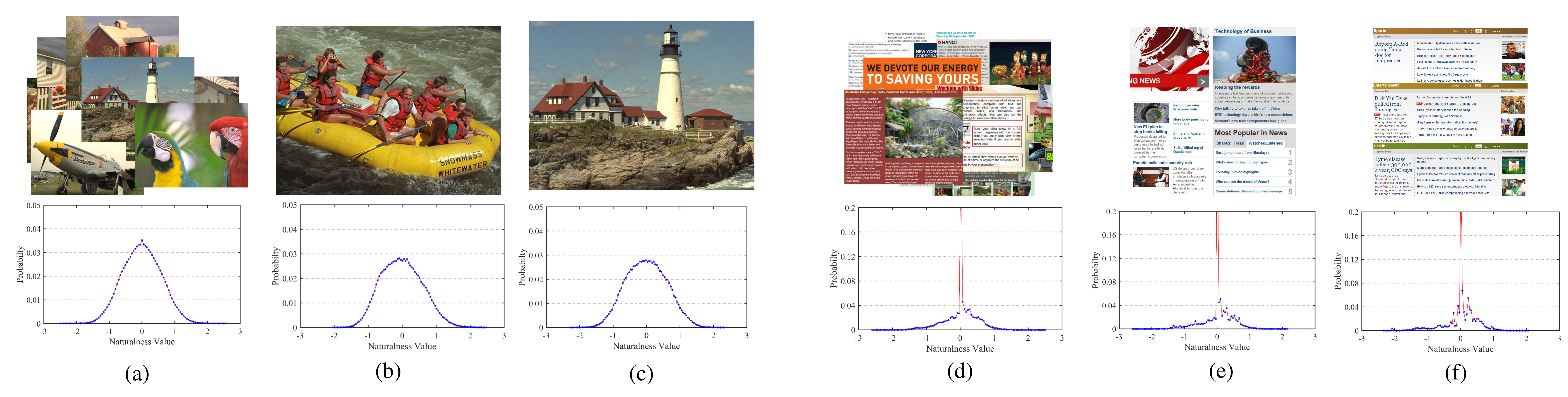}}
\end{minipage}
\caption{\bl{Distribution of Naturalness Values (DNV) of  NIs and SCIs. (a) DNV of all reference images in TID2013; (b) and (c) DNVs of two sample images in TID2013 database; (d)  DNV of all reference images in SIQAD; (e) and (f) DNVs of two sample images in SIQAD database. }}
\label{stat}
\end{figure*}
\subsection{Domain Adaptation} Transfer learning has emerged as an effective technique to address the lack of sufficient labels on target domain by learning from labelled source domain. However, the domain gap between source and target domains is inclined to cause the performance degradation. To alleviate this issue, DA has been widely concerned. Common practices are matching the feature distributions in the source and the target domains and minimizing the feature discrepancy. Some approaches achieve this target by reweighing or selecting samples from the source domain \cite{borgwardt2006integrating,huang2007correcting,gong2013connecting}, while others focus on explicit feature space transformation and attempt to map distributions of two domains by minimizing the maximum mean discrepancy (MMD) \cite{long2017deep}\cite{yan2017mind}, correlation alignment (CORAL) \cite{sun2016deep}\cite{peng2018synthetic}, or reducing the Kullback-Leibler (KL) divergence \cite{zhuang2015supervised}.  Considering the label difference between source domain and target domain also should be captured in DA, Li \textit{et al.} \cite{li2020unsupervised} proposed to disentangle the latent representation into a global code and a local code, by which the reconstruction of samples can be performed with the categories and non-category style information, leading to a more effective DA method. In \cite{das2018sample, das2018graph,das2018unsupervised}, the graph matching based methods were proposed to encourage the source and target representations to be in support of each other. As such, the model will be further refined by the pseudo labels to shrink the domain discrepancy.

Another line of research is confusing a domain discriminator by adversarial learning. In \cite{ganin2014unsupervised}, a gradient reversal layer is proposed for domain invariant feature learning. Generative adversarial networks (GANs) have also been adopted for DA to transform the appearance of source samples, in an effort to enforce them to be similar with target samples. In ~\cite{taigman2016unsupervised,bousmalis2017unsupervised}, the adaptation is achieved by applying adversarial losses in the pixel space for cross-domain image mapping. However, these methods only impose the constraint on the distributions in the feature space while neglecting the distributions of  classification results in the label space, potentially leading to an inadequate DA.

\section{The Proposed Scheme}
\begin{figure*}[t]
\begin{minipage}[b]{0.95\linewidth}
  \centering
  \centerline{\includegraphics[width=1\linewidth]{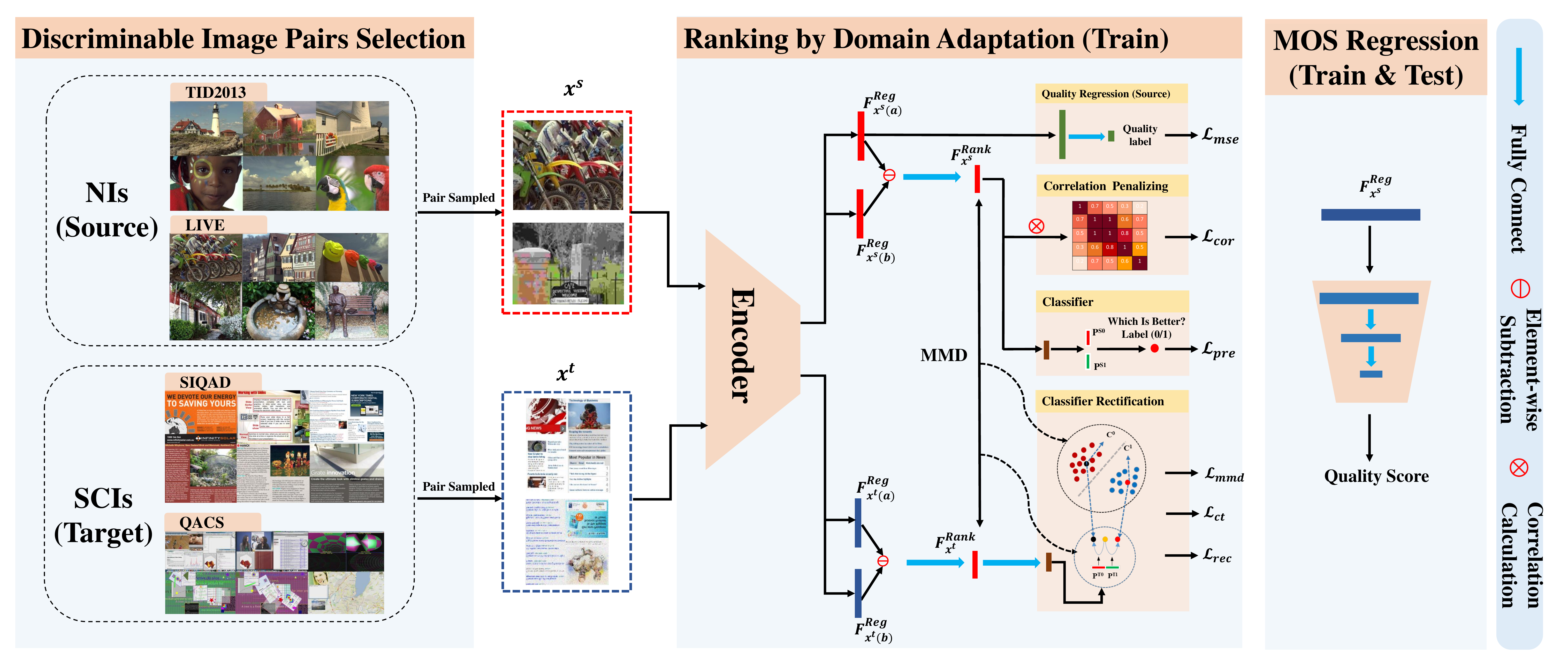}}
\end{minipage}
\caption{ \bl{The framework of the proposed NRIQA for SCIs with UDA. }}
\label{framework}
\end{figure*}
\subsection{Motivation}
Our goal is to learn an NRIQA model that captures and quantifies the regularities of SCIs based on ground-truth ratings from NIs. Instead of simply transferring the continuous quality ratings, we learn to rank the quality of image pairs, which has been regarded as an alternative yet promising paradigm in conveying the quality, but has not been fully exploited in DA based quality prediction task.   
It is generally acknowledged that the statistics of naturalness for SCIs are dramatically different from those of NIs. In this regard, one empirical experiment is conducted. In particular, we compute the naturalness distributions of NIs and SCIs following \cite{mittal2012making}, 
\begin{equation}
\psi(i, j)=\frac{I(i, j)-\mu(i, j)}{\sigma(i, j)+1},
\end{equation}
where $i$ and $j$ are the spatial indices in an image $I$. The mean $\mu(i,j)$ and deviation $\sigma (i,j)$ are computed as follows,
\begin{equation}
\mu(i, j)=\sum_{k=-K}^{K} \sum_{l=-L}^{L} \omega_{k, l} I(i+k, j+l),
\end{equation}
\begin{equation}
\sigma(i, j)=\sqrt{\sum_{k=-K}^{K} \sum_{l=-L}^{L} \omega_{k, l}[I(i+k, j+l)-\mu(i, j)]^{2}},
\end{equation}
where $\omega$ is the Gaussian weighting function with $K = L = 3$. \bl{As shown in Fig.~\ref{stat}, we show the distributions of naturalness values of both NIs in TID2013 database~\cite{ponomarenko2015image} and SCIs in SIQAD database~\cite{yang2015perceptual}. It is apparent that the distribution of NIs is in stark contrast compared to the statistical distribution of SCIs, due to the fact that SCIs are computer generated instead of being acquired by optical cameras.} 
Inspired by such statistics gap, we transform the quality prediction with UDA to a relatively less ambiguous task of {inferring the quality ranking of image pairs}. This aligns with the cognitive process, as it is usually much more straightforward to compare a pair of images than providing the rating scales (e.g., 5-star). 
Second, learning from image pairs can highly expand the training samples which further alleviates the over-fitting problem to some extent. Third, it is quite feasible to obtain the global quality predictions by aggregating from pairwise comparison data. These motivations inspire us to explore the transferability of pairwise relationship between the source and target domains. 
\subsection{Framework}
The framework of the proposed method is illustrated in Fig.~\ref{framework}. In the training stage, our ultimate goal is to learn the encoder with a tailored SCNN \cite{zhang2018blind} as the backbone, due to its light-weight and high efficiency. {The encoder aims
to generate the quality deterministic feature \textit{$\boldsymbol{F}^{Reg}$}, based on which a quality regressor that is able to predict the quality in the target domain can be learnt. To this end, we first enforce \textit{$\boldsymbol{F}^{Reg}$} to be able to predict the ground truth quality in the source domain with a quality regression module, and more importantly, \textit{$\boldsymbol{F}^{Reg}$} should also be able to convey the quality ranking in both source and target domains. In the pairwise quality learning, the ranking feature is obtained via the element-wise comparisons between the quality features from a pairs of images elaborately selected.} Generally speaking, three constrains are imposed on the quality ranking feature \textit{$\boldsymbol{F}^{Rank}$}, including {1) discriminantive: the features should be able to accurately predict the ranking label in the source domain; 2) transferable: the feature discrepancy between the two domains are shrunk with the MMD loss and the binary ranking relation of the two domains are unified by two aligned clusters with the center based rectification; 3) inclusive:
the features are enriched by reducing the correlations between different dimensions in the feature space, embodying quality deterministic factors and leading to the features with high generalization capability~\cite{bengio2013representation}.} Given \textit{$\boldsymbol{F}^{Reg}$}, to learn the
quality regression network in the target domain, the pseudo labels which are obtained based on the rankings of SCIs are generated for training. In the testing stage, given one image as the input to the encoder, the \textit{$\boldsymbol{F}^{Reg}$} is generated based on
which the regression network can produce the final quality.


\subsection{Quality Deterministic Feature Generation}


For feature generator, we use a pretrained CNN performed on the selected  discriminable image pairs to extract feature of each single image. The CNN adopted is the tailored SCNN introduced in \cite{zhang2018blind}, which is pretrained with different distortion types on NIs. Here, only the second-to-last layers are used as the feature encoder. There are two reasons behind which we choose SCNN as our 
encoder. First, the SCNN is a much lighter network compared with other pretrained networks such as VGG \cite{simonyan2014very} or ResNet \cite{he2016deep}. Second, the prior knowledge of distortion types that SCNN obtains could make our model more stable during training, as evidenced by our experiments. As described in above section, given an NI image pair \(x\) (source domain) which consists of two images  \(x (a)\) and \(x(b)\), we utilize the first image  \(x(a)\)  to go through the  feature extractor  SCNN and the  result is denoted as \textit{$\boldsymbol{F}^{Reg}_{x(a)}$}.
Then a constrained fully connect layer (the weights of which are constrained to be positive) is designed to infer the image quality as follows,
\begin{equation}\label{reg}
\begin{aligned}
\textit{${P}^{Reg}_{x(a)}$} = FC(\textit{$\boldsymbol{F}^{Reg}_{x(a)}$}),
\end{aligned}
\end{equation}
where \(FC\) is the fully connect layer and its output dimension is 1. \textit{${P}^{Reg}_{x(a)}$}  is the predicted score of image \(x(a)\) and we use the MSE (Mean Square Error) loss  for quality regression:
\begin{equation}\label{mse2}
\mathcal{L}_{mse}= \frac{1}{n}
\sum_{i=1}^{n}(T_{x(a)}- \textit{${P}^{Reg}_{x(a)}$} )^2,
\end{equation}
where the \(T_{x(a)}\) is the ground truth quality of \(x(a)\) and \(n\) is the number of NIs in a batch.   The first challenge in this new setting is the unavailable label information in the target domain. Traditional image quality prediction models, which take advantage of the neural network aiming to approach the ground truth MOS, can only ensure the quality model with the capability of extracting quality deterministic features in source domain. To confront this challenge, pair-wise quality learning is further proposed, together with which the quality of source domain can be well transferred to the target domain. 
{Before the pair-wise quality learning performed, we denote the ranking feature of an image pair \(x\)  as \textit{$\boldsymbol{F}^{Rank}_{x}$} which can be estimated by measuring the differences of the corresponding features \textit{$\boldsymbol{F}^{Reg}_{x(a)}$} and \textit{$\boldsymbol{F}^{Reg}_{x(b)}$} as follows,
\begin{equation}\label{}
\textit{$\boldsymbol{F}^{Rank}_{x}$} = \textit{$\boldsymbol{F}^{Reg}_{x(a)}$} - \textit{$\boldsymbol{F}^{Reg}_{x(b)}$}.
\end{equation}
}

\subsection{Pair-wise Quality Learning} 
The pairwise quality learning aims to equip the quality features with the capability to accurately predict the ranking in both source and target domains, though the label information in the target domain is unavailable. We first introduce the preliminary, following which the loss functions are defined. Finally, we discuss the pair selection strategy, in an effort to carefully select the discriminative pairs in the training stage. 

\subsubsection{Preliminary}
Given $n^{s}$ NI pairs ${x}^{s}$ and their labels ${y}^{s} = \{0,1\}$ in the NI dataset as source domain $\mathcal{D^S}$: $\left\{\left({x_i}^{s},y_i^{s}\right)\right\}_{i=1}^{n^{s}}$ and $n^{t}$ unlabelled SCI pairs  ${x}^{t}$ as target domain $\mathcal{D^T}$:$\left\{{x_j}^{t}\right\}_{j=1}^{n^{t}}$, our task is to learn a feature extractor $G(\cdot)$ and a ranking classifier $C(\cdot)$ such that the expected target risk $\mathcal{E}_{\left({x_i}^{t}, y_i^{t}\right) \sim \mathcal{D^T}}\left[\mathcal{L}_{c l s}\left(C\left(G\left({x_i}^{t}\right)\right), y_i^{t}\right)\right]$ can be minimized with a certain classification loss function $\mathcal{L}_{cls}(\cdot)$, where $y_i^t$ denotes the corresponding ground truth of ${x_i}^t$ from target domain. To be specific, we denote two images as $x_{i}(a)$ and $x_{i}(b)$ in an image pair $x_{i}$, assuming their quality values are  $Q_{a}$ and $Q_{b}$, then the probability of the ranking classification $P_{x_{i}}^{rank}$ can be estimated as follows:
\begin{equation}\label{xx}
\begin{aligned}
P_{x_{i}}^{rank} &= P((y_{i}=1) |x_{i})\\ 
&= P\left((Q_a > Q_b)| (x_{i}(a) , x_{i}(b))\right),
\end{aligned}
\end{equation}
where \(P(\cdot)\) represents the probability. In \cite{ben2007analysis,ben2010theory}, Ben-David \textit{et al.} proved that the upper bound of the empirical risk on the target domain is jointly determined by the empirical risk on the source domain and the discrepancy between the source and target domains. 
In this paper, we model the distance between source and target domains by considering the joint distribution of data pair, which can be formulated as follows:
\begin{equation}\label{eqdis}
\mathcal{DIST}\left(\mathcal{D}^{s}, \mathcal{D}^{t}\right) =  d(p\left({x}^{s},{y}^{s}\right), p\left({x}^{t},{y}^{t}\right)),
\end{equation}
where \(d(\cdot,\cdot)\) is a probability discrepancy measurement. Moreover,  $p\left({x}^{s},{y}^{s}\right)$ and $p\left({x}^{t},{y}^{t}\right)$ are the joint distributions which can be modelled by the marginal probability distribution  and conditional probability distribution,
\begin{equation}\label{eqmbyn}
\begin{aligned}
p\left({x}^{s},{y}^{s}\right) = p\left({x}^{s}\right)   p\left({y}^{s} |  {x}^{s}\right),  \\
p\left({x}^{t},{y}^{t}\right) = p\left({x}^{t}\right)  p\left({y}^{t} |   {x}^{t}\right),
\end{aligned}
\end{equation}
where $p\left({x}^{s}\right)$ and $p\left({x}^{t}\right)$ denote the marginal distribution on the ranking features of source and target domains, respectively. Moreover, $p\left({y}^{s} |   {x}^{s}\right)$ and  $p\left({y}^{t} |   {x}^{t}\right)$ denote the conditional probability distribution on the ranking outputs conditioned on the ranking features of source and target domains, respectively. 
The empirical risk on the source domain can be minimized with the cross-entropy loss for the purpose of pairwise ranking classification based on the labelled NI pairs.
To further minimize the empirical risk of the target domain, we propose to reduce the domain distribution discrepancy by jointly constraining the distance between
$ p\left({x}^{s}\right)$ and $ p\left({x}^{t}\right)$ in the feature space by a MMD loss and the distance between  $p\left({y}^{s} |  {x}^{s}\right)$ and $p\left({y}^{t} |  {x}^{t}\right)$  by a classifier rectification loss.

\subsubsection{Empirical risk minimization on source domain}
{
For the ranking feature classification, a classifier which consists of a fully connect ($FC$) layer and a \(softmax\) layer is adopted as follows:
\begin{equation}
\begin{aligned}
\textit{${P}^{Rank}_{x^s_i}$} = Softmax(FC(\textit{$\boldsymbol{F}^{Rank}_{x^s_i}$})),
\end{aligned}
\end{equation}
where \textit{${P}^{Rank}_{x^s_i}$} represents the predicted probability that the first image has better quality than the second one in the $i-th$ input NI pair of a batch. Then we adopt the binary cross entropy loss for the network training:
\begin{equation}
\begin{aligned}
\mathcal{L}_{\text {pre}}=\frac{1}{N} \sum_{i=1}^{N}-\left[y^s_{i} \cdot \log \textit{${P}^{Rank}_{x^s_i}$} +\left(1-y^s_{i}\right) \cdot \log  \textit{${P}^{Rank}_{x^s_i}$} \right],
\end{aligned}
\end{equation}
where $N$ is is the size of NI pairs in a batch, and $i$ indicates the $i-th$ input pair with its binary label denoted as $y^s_{i}$. 

{To further mitigate the over-fitting of \textit{$\boldsymbol{F}^{Rank}$}, we encourage to involve quality deterministic factors in \textit{$\boldsymbol{F}^{Rank}$}~\cite{bengio2013representation}.} Inspired by the works in \cite{breiman1996bagging,cogswell2015reducing,rodriguez2016regularizing}, we adopt the correlation penalization on \textit{$\boldsymbol{F}^{Rank}$} in each batch, by which low-redundant ranking feature is expected to be learnt from the source domain. Assuming in a batch, \textit{$\boldsymbol{F}^{Rank}$}  $\in \mathbb{R}^{N \times d} $, where $N$ is the size of NI pairs in a batch and $d$ is the length/dimension, then the correlation metric \textit{$\boldsymbol{C}$}  (with size of  $d \times d$) can be formed by calculating the correlation between all dimension pairs of \textit{$\boldsymbol{F}^{Rank}$}: 
\begin{equation}\label{}
\textit{$\boldsymbol{C}_{m,n}$} = \frac{(\textit{$\boldsymbol{F}^{Rank}_{x^s_{all},d_m}$}-\mu_m \textit{$\boldsymbol{1}$})^T(\textit{$\boldsymbol{F}^{Rank}_{x^s_{all},d_n}$}-\mu_n \textit{$\boldsymbol{1}$})}
{\sqrt{\sum_{i=1}^{N}(\textit{$\boldsymbol{F}^{Rank}_{x^s_i,d_m}$}-\mu_m)^2}
\sqrt{\sum_{i=1}^{N}(\textit{$\boldsymbol{F}^{Rank}_{x^s_i,d_n}$}-\mu_n)^2}},
\end{equation}
where $\textit{$\boldsymbol{F}^{Rank}_{x^s_{all},d_m}$}$  represents the concatenating of the $m-th$ dimension elements of $\textit{$\boldsymbol{F}^{Rank}$}$ of all NI pairs in a batch. $\textit{$\boldsymbol{F}^{Rank}_{x_{all},d_n}$}$ has the same meaning of $\textit{$\boldsymbol{F}^{Rank}_{x_{all},d_m}$}$ for the $n-th$ dimension selected. The sizes of  $\textit{$\boldsymbol{F}^{Rank}_{x_{all},d_m}$}$ and $\textit{$\boldsymbol{F}^{Rank}_{x_{all},d_n}$}$ are all $N \times 1$. Moreover, $\mu_m$ and $\mu_n$ are the mean values of $\textit{$\boldsymbol{F}^{Rank}_{x_{all},d_m}$}$ and  $\textit{$\boldsymbol{F}^{Rank}_{x_{all},d_n}$}$, and  $\textit{$\boldsymbol{1}$}$ is a vector the dimension of which is identical with $\textit{$\boldsymbol{F}^{Rank}_{x_{all},d_m}$}$ with all elements equaling to 1. Then we penalize the norm of $\boldsymbol{C}$ by loss $\mathcal{L}_{cor}$ as follows:
\begin{equation}\label{}
\mathcal{L}_{cor} = \frac{1}{d \times d}\left\|\textit{$\boldsymbol{C}$}-\textit{$\boldsymbol{I}$}\right\|_2,
\end{equation}
where  $\left\|\cdot \right\|_2$ is the square normal and  \textit{$\boldsymbol{I}$} is the identity matrix of the same dimension as \textit{$\boldsymbol{C}$}. We subtract \textit{$\boldsymbol{I}$} here due to the diagonal of \textit{$\boldsymbol{C}$} is a constant vector (all elements are equaling to 1) such that its gradient does not need to be calculated during training.

}

\subsubsection{Empirical risk minimization on target domain with domain alignment}
Herein, we introduce how the source and target domains could be aligned to transfer the quality from the source to the target domain. Though it is generally acknowledged that there are dramatical differences in terms of the statistics presumably perceived by HVS between source and target domains, the shareable feature responses, which are transferable and subjected to be learnt with DA,  originate from the relatively quality rank across content and even distortion types.
As discussed above, we propose to conduct domain adaptation by jointly considering the marginal distribution on ranking features, and the ranking output distribution conditioned on the features. 

Regarding the marginal distribution based on the ranking features, we propose to reduce the discrepancy of \textit{$\boldsymbol{F}^{Rank}$} in source domain and target domain by a Maximum Mean Discrepancy (MMD) loss, which can be formulated as 
\begin{equation}\label{}
\mathcal{L}_{m m d}=\left\|\frac{1}{N} \sum_{i=1}^{N} \phi\left(\boldsymbol{F}^{Rank}_{x^s_i}\right)-\frac{1}{K} \sum_{j=1}^{K} \phi\left(\boldsymbol{F}^{Rank}_{x^t_j}\right)\right\|_{\mathcal{H}}^{2},
\end{equation}
where the $N$ and $K$ indicate the numbers of NI pairs and SCI pairs in a batch respectively and $\phi$ is a function that maps the features into the Reproducing kernel Hilbert Space (RKHS) \cite{gretton2012kernel}.  We apply the Kernel trick by adopting Gaussian kernel \cite{gretton2012kernel} to compute $\mathcal{L}_{m m d}$ by setting the kernel bandwidth to be the median distances of all pairwise data points from the batch, such that the discrepancy of marginal distribution is expected to be minimized. 

Regarding the conditional probability distribution based on ranking features and the corresponding ranking outputs, we propose a classifier rectification loss to improve the discrimination capability of classifier on target domain. More concretely, we first apply a center loss \cite{wen2016discriminative} to learn two centers for the ranking feature of each class on the source domain as follows: 
\begin{equation}\label{ctt1}
\begin{split}
\mathcal{L}_{ct}= \frac{1}{N}\sum_{i=1}^{N}&\left(\delta\left(y^s_{i}=0\right)\left\|\boldsymbol{F}^{Rank}_{x^s_i}-\boldsymbol{c}^{0}\right\|_{2}^{2} \right.\\  
& \left.+\delta\left(y^s_{i}=1\right)\left\|\boldsymbol{F}^{Rank}_{x^s_i}-\boldsymbol{c}^{1}\right\|_{2}^{2} \right),
\end{split}
\end{equation}
where $\delta(condition) = 1$ if the condition is satisfied, and $\delta(condition) = 0$ otherwise. In addition, $\boldsymbol{c^0}$ and $\boldsymbol{c^1}$ are the learned class specific centers. The center loss simultaneously learns the centers of each class and penalizes the distances between the features. In particular, with the centers acquired from the source domain, they can be further applied to cluster the ranking features in the target domain,
\begin{equation}\label{ctt}
\begin{split}
\mathcal{L}_{rec}= \frac{1}{K}\sum_{j=1}^{K}&\left((1-\textit{${P}^{Rank}_{x^t_j}$})\left\|\boldsymbol{F}^{Rank}_{x^t_j}-\boldsymbol{c}^{0}\right\|_{2}^{2}\right.\\  
&  \left.+\textit{${P}^{Rank}_{x^t_j}$}\left\|\boldsymbol{F}^{Rank}_{x^t_j}-\boldsymbol{c}^{1}\right\|_{2}^{2}\right),
\end{split}
\end{equation}
where \textit{${P}^{Rank}_{x^t_j}$} indicates the likelihood of the $j-th$ sample be classified to class 1. There are two advantages when imposing $\mathcal{L}_{rec}$. First, the wrong classification results will be rectified based on the distance between features and the two centers in the target domain, as the $\mathcal{L}_{rec}$ tends to decrease when the ranking features are classified to its closest center with high probability. Second, the center of each class can be updated gradually when the ranking features of NIs and SCIs vary. This could further improve the feature separability, and finally the two class-specific centers shared by NIs and SCIs can be acquired.

\subsubsection{Discriminable image pairs selection}

One may consider to generate rankings with arbitrary images from source and target domains, in an effort to create the training set for ranking. However, we argue that this may not be optimal due to two reasons. First, the  intrinsically ambiguous pairs are even difficult to be distinguished by HVS, particularly when the quality scores of the image pair are extremely close. As such, the labels of such hard sample pairs may not be credible \cite{ma2017dipiq} and cause the difficulty in model convergence when they are involved during training. Second, forcing the network to distinguish the indiscriminable image pairs is likely to lead to overfitting problem to source domain, resulting in a negative transferring to the target domain. To this end, we propose to select the image (both NI and SCI) pairs 
with their quality score difference governed by a threshold, rendering \textit{discriminable pairs} instead of random pairs.


{As the ground truth information is not available in the target domain, we propose to use the DB-CNN model \cite{zhang2018blind}, which is pre-trained on NI dataset (TID2013), to predict the pseudo ratings of SCIs as the guidance. It is worth mentioning that although the predicted quality may not be accurate enough,  they only serve as the guidance based on the difference between image pairs, such that the discriminable image pairs can be selected by a predefined threshold. In our method, we also find the selection process is also necessary for our classifier rectification, 
making our model easier to learn the center features shared by NIs and SCIs.}

\subsection{Quality Prediction}
The total loss functions can be summarized as follows,
\begin{equation}\label{all}
\mathcal{L}=\mathcal{L}_{\text{pre}} +  \lambda_0 \mathcal{L}_{m m d} + \lambda_1 \mathcal{L}_{ct} + \lambda_2 \mathcal{L}_{rec}+\lambda_3 \mathcal{L}_{cor}+\lambda_4 \mathcal{L}_{mse},
\end{equation}
where $ \lambda_{0} $, $ \lambda_{1} $, $ \lambda_{2} $, $ \lambda_{3} $, $ \lambda_{4} $ are the weighting factors. Given the trained ranking model, it is further applied to predict the quality of SCIs. To precisely obtain the quality, for an image $I_i$, we compare it with the rest of all of images and its quality score $S_{I_i}$ is given by,
\begin{equation}\label{mos}
S_{I_i} = \frac{\sum_{j=1}^{n^t-1}\delta\left[P\left((Q_i > Q_j)| (I_{i} , I_{j})\right)>0.5 \right]}{n^t-1},
\end{equation}
where $n^t$ is the number of all the SCIs, $P\left((Q_i > Q_j)| (I_{i} , I_{j})\right)$ is the ranking  classification result and  $\delta(condition) = 1$ if the condition is satisfied, and $\delta(condition) = 0$ otherwise. After obtaining the predicted quality score of all SCIs, those scores will be treated as pseudo labels of all samples in the target domain. Therefore, a quality regression network can be trained on the target domain. More specifically, we fix the ranking model parameters and use the extracted distortion features $\boldsymbol{F}^{Reg}$ as input, followed by two fully connect layers. Then the regression network is trained with the following loss,
\begin{equation}\label{mse}
\mathcal{L}_{mse}^t= \frac{1}{L}
\sum_{i=1}^{n^t}(\textit{${S}^{Reg}_{I_i}$}-S_{I_i})^2,
\end{equation}
where the $I_i$ is the $i-{th}$ sample in a batch and $S^{Reg}_{I_i}$ is the predicted result by the regression network. In our experiments, we find the regression network can reduce the noise in the pseudo labels and further enhance the accuracy of quality prediction. With the refined pseudo labels, the discriminable image pairs re-selection can be achieved instead of relying on the pre-trained DB-CNN model only, such that our model can be further refined by retraining.
\begin{table*}
  \centering
  \caption{\bl{Description of the NI and SCI databases.}}
\setlength{\tabcolsep}{1.8mm}{
\begin{tabular}{c|c|c|c|c}
\hline
\multicolumn{2}{c|}{\textbf{Database}} & \multicolumn{1}{m{3.79em}<{\centering}|}{\textbf{\# of Ref. Images}} &
\multicolumn{1}{m{3.79em}<{\centering}|}{\textbf{\# of Images}}&
\textbf{Distortion Type} \\

\hline
\multicolumn{1}{c|}{\multirow{2}[4]{*}{\shortstack{Source \\domain}}} & TID2013 & 25    & 3000  & \multicolumn{1}{m{29.79em}<{\centering}}{(1) Additive Gaussian noise; (2) Additive noise in color components; (3) Spatially  correlated noise;   (4) Masked noise; (5) High frequency noise; (6) Impulse noise; (7) Quantization noise; (8) Gaussian blur; (9) Image denoising; (10) JPEG compression; (11) JPEG2000 compression;  (12) JPEG transmission errors; (13) JPEG2000 transmission errors; (14) Non eccentricity pattern noise; (15) Local block-wise  distortions of different intensity; (16) Mean shift (intensity shift); (17) Contrast change; (18) Change of color saturation; (19) Multiplicative Gaussian noise; 
(20) Comfort noise; (21) Lossy compression of noisy images; (22) Image color 
quantization with dither; (23) Chromatic aberrations; (24) Sparse sampling and reconstruction} \\
\cline{2-5}      & LIVE  & 29    & 982   & \multicolumn{1}{m{29.79em}<{\centering}}{(1) JPEG compression (2) JPEG2000 compression; (3) Additive white Gaussian noise (4) Gaussian blur; (5) Rayleigh fast-fading channel distortion}   \\
\hline
\multicolumn{1}{c|}{\multirow{2}[4]{*}{\shortstack{Target \\domain}}} & SIQAD & 20    & 980   &  \multicolumn{1}{m{29.79em}<{\centering}}{(1) Gaussian noise; (2) Gaussian blur; (3) Motion blur; (4) Contrast change; (5) JPEG; (6) JPEG2000; (7) Layer Segmentation based Coding} \\
\cline{2-5}      & QACS  & 24    & 492   &  \multicolumn{1}{m{29.79em}<{\centering}}{(1) HEVC Compression; (2) SCC Compression}  \\
\hline
\end{tabular}%

}
 \label{tab:list}%
\end{table*}%

\begin{table}
  \centering
  \caption{Architecture of the network in the proposed method.}
\setlength{\tabcolsep}{1mm}{
\begin{tabular}{cc|cccc}
\hline
\hline
\multicolumn{2}{c|}{\textbf{Layer Type}} & \multicolumn{1}{c|}{\textbf{Kernel Size}} & \multicolumn{1}{c|}{\textbf{Channels}} & \multicolumn{1}{c|}{\textbf{Stride}} & \textbf{Padding} \bigstrut\\
\hline
\hline
\multicolumn{1}{c|}{\multirow{16}[30]{*}{\textbf{\shortstack{SCNN \\\cite{zhang2018blind}}}}} & \textbf{conv} & 3x3   & [3,48] & 1     & 1 \bigstrut\\
\cline{2-6}\multicolumn{1}{c|}{} & \multicolumn{5}{c}{\textbf{BN, ReLu}} \bigstrut\\
\cline{2-6}\multicolumn{1}{c|}{} & \textbf{conv} & 3x3   & [48,48] & 2     & 1 \bigstrut\\
\cline{2-6}\multicolumn{1}{c|}{} & \multicolumn{5}{c}{\textbf{BN, ReLu}} \bigstrut\\
\cline{2-6}\multicolumn{1}{c|}{} & \textbf{conv} & 3x3   & [48,64] & 2     & 1 \bigstrut\\
\cline{2-6}\multicolumn{1}{c|}{} & \multicolumn{5}{c}{\textbf{BN, ReLu}} \bigstrut\\
\cline{2-6}\multicolumn{1}{c|}{} & \textbf{conv} & 3x3   & [64,64] & 2     & 1 \bigstrut\\
\cline{2-6}\multicolumn{1}{c|}{} & \multicolumn{5}{c}{\textbf{BN, ReLu}} \bigstrut\\
\cline{2-6}\multicolumn{1}{c|}{} & \textbf{conv} & 3x3   & [64,64] & 1     & 1 \bigstrut\\
\cline{2-6}\multicolumn{1}{c|}{} & \multicolumn{5}{c}{\textbf{BN, ReLu}} \bigstrut\\
\cline{2-6}\multicolumn{1}{c|}{} & \textbf{conv} & 3x3   & [64,64] & 2     & 1 \bigstrut\\
\cline{2-6}\multicolumn{1}{c|}{} & \multicolumn{5}{c}{\textbf{BN, ReLu}} \bigstrut\\
\cline{2-6}\multicolumn{1}{c|}{} & \textbf{conv} & 3x3   & [64,128] & 1     & 1 \bigstrut\\
\cline{2-6}\multicolumn{1}{c|}{} & \multicolumn{5}{c}{\textbf{Average Pooling  14 × 14}} \bigstrut\\
\cline{2-6}\multicolumn{1}{c|}{} & \textbf{FC} & \multicolumn{4}{c}{[128,256]} \bigstrut[t]\\
\multicolumn{1}{c|}{} & \textbf{FC} & \multicolumn{4}{c}{[256,256]} \bigstrut[b]\\
\hline
\multicolumn{2}{c|}{\textbf{FC}} & \multicolumn{4}{c}{[256,128]} \bigstrut\\
\hline
\multicolumn{3}{c|}{\textbf{Quality Regression (Source)}} & \multicolumn{3}{c}{\textbf{Subtraction Layer}} \bigstrut\\
\hline
\multicolumn{2}{c|}{\textbf{FC}} & \multicolumn{1}{c|}{[128,1]} & \multicolumn{1}{c|}{\textbf{FC}} & \multicolumn{2}{c}{[128,2]} \bigstrut\\
\hline
\hline
\multicolumn{6}{c}{\textbf{Quality Regression (Target)}} \bigstrut\\
\hline
\hline
\multicolumn{2}{c|}{\textbf{FC}} & \multicolumn{4}{c}{[128,256]} \bigstrut[t]\\
\multicolumn{2}{c|}{\textbf{FC}} & \multicolumn{4}{c}{[256,1]} \bigstrut[b]\\
\hline
\end{tabular}%

}
  \label{tab:network}%
\end{table}%

\section{Experimental Results}
\label{sec:typestyle}

\begin{table*}
  \centering
  \caption{Quality prediction performance comparisons (target domain: SIQAD).}
\begin{tabular}{c|c|ccccc}
\hline
\multicolumn{2}{c|}{Target domain: SIQAD} & PLCC  & MAE   & RMSE   & SRCC  & KRCC \bigstrut\\
\hline
\multirow{4}[2]{*}{Conventional NIQA} & NIQE  & 0.2967  & 11.0587  & 13.5484  & 0.2863  & 0.1963  \bigstrut[t]\\
      & PIQE  & 0.4588  & 10.2610  & 12.6062  & 0.3302  & 0.2264  \\
      & SSEQ  & 0.4758  & 10.1789  & 12.4787  & 0.3467  & 0.2380  \\
      & BRISQ & 0.4474  & 10.2684  & 12.6879  & 0.4159  & 0.2833  \\
      & HOSA   & 0.2226  & 11.5618  & 13.9550  & 0.2194  & 0.1451   \\
& BMPRI  &0.3393   &10.7864   &13.4652    &0.3208    &0.2328 \\
& BPRI   &0.5862    &8.9985   &11.5972    &0.5564    &0.4010  \\
\hline
\multicolumn{1}{c|}{\multirow{6}[2]{*}{\shortstack{Deep Learning Based \\(Training on TID2013)}}} & Rank  & 0.2547  & 11.2565  & 13.7194  & 0.2303  & 0.1557  \bigstrut[t]\\
      & NRFR (weighted) & 0.2337  & 11.3942  & 13.7943  & 0.0692  & 0.0471  \\
      & NRFR (patch) & 0.4138  & 10.3556  & 12.9156  & 0.3793  & 0.2598  \\
      & MENO  & 0.5384  & 9.5297  & 11.9554  & 0.4350  & 0.3055  \\
      & DB-CNN  & 0.5226  & 9.8784  & 12.0960  & 0.5294  & 0.3652  \\
      & Ours  & \textbf{ 0.7117 } & \textbf{ 8.2529 } & \textbf{ 9.9663 } & \textbf{ 0.6970 } & \textbf{ 0.4895 } \bigstrut[b]\\
\hline
\multicolumn{1}{c|}{\multirow{5}[2]{*}{\shortstack{Deep Learning Based \\(Training on LIVE)}}} & Rank  & 0.1745  & 11.4861  & 13.9697  & 0.1695  & 0.1149  \bigstrut[t]\\
      & NRFR (weighted) & 0.4705  & 10.0769  & 12.5189  & 0.4655  & 0.3227  \\
      & NRFR (patch) & 0.4569  & 10.0070  & 12.6199  & 0.4536  & 0.3237  \\
      & DB-CNN  & 0.5629  & 9.4289  & 11.7266  & 0.5055  & 0.3237  \\
      & Ours  & \textbf{ 0.7026 } & \textbf{ 8.3833 } & \textbf{ 10.0954 } & \textbf{ 0.6904 } & \textbf{ 0.4880 } \bigstrut[b]\\
\hline
\end{tabular}%

  \label{tab:siq}%

\end{table*}%

\begin{table*}
  \centering
  \caption{Quality prediction performance with different distortion types  (target domain: SIQAD).}
\begin{tabular}{cc|rr|rr|rr|rr}
\hline
\multicolumn{2}{c|}{\multirow{6}[12]{*}{\shortstack{Training on TID2013   \\Testing on SIQAD}}} & \multicolumn{2}{c|}{JPEG} & \multicolumn{2}{c|}{ Gaussian Noise} & \multicolumn{2}{c|}{ Gaussian Blur} & \multicolumn{2}{c}{ JPEG2000 } \bigstrut\\
\cline{3-10}\multicolumn{2}{c|}{} & \multicolumn{1}{c}{PLCC} & \multicolumn{1}{c|}{SRCC} & \multicolumn{1}{c}{PLCC} & \multicolumn{1}{c|}{SRCC} & \multicolumn{1}{c}{PLCC} & \multicolumn{1}{c|}{SRCC} & \multicolumn{1}{c}{PLCC} & \multicolumn{1}{c}{SRCC} \bigstrut\\
\cline{3-10}\multicolumn{2}{c|}{} & 0.7883  & 0.5053  & 0.2657  & 0.1512  & 0.5908  & 0.5207  & 0.6722  & 0.5541  \bigstrut\\
\cline{3-10}\multicolumn{2}{c|}{} & \multicolumn{2}{c|}{ Contrast Change} & \multicolumn{2}{c|}{LSC} & \multicolumn{2}{c|}{ Motion Blur} & \multicolumn{2}{c}{All} \bigstrut\\
\cline{3-10}\multicolumn{2}{c|}{} & \multicolumn{1}{c}{PLCC} & \multicolumn{1}{c|}{SRCC} & \multicolumn{1}{c}{PLCC} & \multicolumn{1}{c|}{SRCC} & \multicolumn{1}{c}{PLCC} & \multicolumn{1}{c|}{SRCC} & \multicolumn{1}{c}{PLCC} & \multicolumn{1}{c}{SRCC} \bigstrut\\
\cline{3-10}\multicolumn{2}{c|}{} & 0.6511  & 0.5563  & 0.7372  & 0.6289  & 0.7735  & 0.6633  & 0.7117  & 0.6970  \bigstrut\\
\hline
\hline
~\\
\hline
\hline
\multicolumn{2}{c|}{\multirow{6}[12]{*}{\shortstack{Training on LIVE   \\Testing on SIQAD}}} & \multicolumn{2}{c|}{JPEG} & \multicolumn{2}{c|}{ Gaussian Noise} & \multicolumn{2}{c|}{ Gaussian Blur} & \multicolumn{2}{c}{ JPEG2000 } \bigstrut\\
\cline{3-10}\multicolumn{2}{c|}{} & \multicolumn{1}{c}{PLCC} & \multicolumn{1}{c|}{SRCC} & \multicolumn{1}{c}{PLCC} & \multicolumn{1}{c|}{SRCC} & \multicolumn{1}{c}{PLCC} & \multicolumn{1}{c|}{SRCC} & \multicolumn{1}{c}{PLCC} & \multicolumn{1}{c}{SRCC} \bigstrut\\
\cline{3-10}\multicolumn{2}{c|}{} & 0.7102  & 0.5023  & 0.3088  & 0.1672  & 0.5734  & 0.5080  & 0.6408  & 0.6034  \bigstrut\\
\cline{3-10}\multicolumn{2}{c|}{} & \multicolumn{2}{c|}{ Contrast Change} & \multicolumn{2}{c|}{LSC} & \multicolumn{2}{c|}{ Motion Blur} & \multicolumn{2}{c}{All} \bigstrut\\
\cline{3-10}\multicolumn{2}{c|}{} & \multicolumn{1}{c}{PLCC} & \multicolumn{1}{c|}{SRCC} & \multicolumn{1}{c}{PLCC} & \multicolumn{1}{c|}{SRCC} & \multicolumn{1}{c}{PLCC} & \multicolumn{1}{c|}{SRCC} & \multicolumn{1}{c}{PLCC} & \multicolumn{1}{c}{SRCC} \bigstrut\\
\cline{3-10}\multicolumn{2}{c|}{} & 0.5898  & 0.5566  & 0.7392  & 0.7193  & 0.7952  & 0.7381  & 0.7026  & 0.6904  \bigstrut\\
\hline
\end{tabular}%
  \label{tab:dist}%

\end{table*}%

\subsection{Experimental Setups}
\subsubsection{Datasets}

To validate the proposed method, we evaluate our model based on four datasets, including two for NIs (TID2013~\cite{ponomarenko2015image} and LIVE~\cite{sheikh2003image}) and two for SCIs (SIQAD~\cite{yang2015perceptual} and QACS~\cite{wang2016subjective}). We adopt four different settings to verify the transferable capability, including: 1) source domain: TID2013, target domain: SIQAD; 2) source domain: LIVE, target domain: SIQAD; 3) source domain: TID2013, target domain: QACS; 4) source domain: LIVE, target domain: QACS. The brief introduction of the dataset is as follows.

\textbf{TID2013.} The TID2013 dataset consists of 3000 images obtained from 25 pristine images for reference. The pristine images are corrupted by 24 distortion types and each distortion type corresponds to 5 levels. 

\textbf{LIVE.} The LIVE IQA database includes 982 distorted NIs and 29 reference images. Five different distortion types are included: JPEG and JPEG2000 compression, additive white Gaussian noise (WN), Gaussian blur (BLUR), and Rayleigh fast-fading channel distortion (FF).

\textbf{SIQAD.} The SIQAD is a SCI dataset which contains 20 source and 980 distorted SCIs. This dataset involves seven distortion types: Gaussian Noise (GN), Gaussian Blur (GB), Motion Blur (MB), Contrast Change (CC), JPEG, JPEG2000 and Layer Segmentation based Coding (LSC). 
Moreover, each distortion type corresponds to seven degradation levels. 

\textbf{QACS.} Compared with SIQAD, QACS database emphasizes on the distortions of compression by two codecs based on the high efﬁciency video coding (HEVC) standard~\cite{sullivan2012overview} and its screen content coding (SCC) extension \cite{shi2015study}. For simplification, the HEVC-SCC extension is denoted as SCC here. This dataset contains 24 source and 492 compressed SCIs. Each SCI is compressed with 11 QP values ranging from 30 to 50, and viewed by twenty subjects with single-stimulus.

\bl{In Table~\ref{tab:list}, the distortion types that are included in the NI database (TID2013 and LIVE) and SCI database (QACS and SIQAD) are presented for better comparison.}

\subsubsection{Implementation Details}
\bl{In the training stage, for discriminable image pairs generation, we first normalize the quality scores of the NI datasets (both TID2013 and LIVE) to the range [0, 1] by min-max normalization strategy. Then the NI pairs are selected based on the criterion that their MOS difference is larger than a given threshold \(\tau=0.07\).}   
By contrast, for SCI pairs, due to the unavailable of the MOS values, we adopt DB-CNN model \cite{zhang2018blind} to estimate their pseudo MOS, which is also normalized to [0, 1] by the min-max normalization strategy. The image pairs of which the MOS difference is larger than a given threshold (0.6 in SIQAD datbase and 0.45 in QACS database) are selected. Besides, the image pairs with the same content (i.e., the structural similarity index (SSIM)~\cite{2004Image} index between the two images is larger than a given threshold 0.75) are also selected to enrich the target domain.


We implement  our  model  by  PyTorch~\cite{paszke2019pytorch}. In Table~\ref{tab:network}, we show the layer-wise network design of our proposed method. The selected images are resized to $224\times 224 \times 3$ as the inputs of our network. The  batch  size  in  the  training  phase  is  16  and  we  adopt Adam optimizer for optimization. The learning rate  is ﬁxed to 5e-5 with a weight decay set as 1e-3. The weighting parameters $ \lambda_{0} $, $ \lambda_{1} $, $ \lambda_{2} $, $ \lambda_{3} $ , $ \lambda_{4} $ in Eqn.~\eqref{all} are set as 1.0, 2e-1 ,1e-3, 1e1 and 1e2 respectively. Early stopping \cite{prechelt1998early} is used when the ranking model is training and the ``patience'' of epochs is set to 2, which implies that the model that provides the best classification accuracy in the source domain is saved until the performance does not improve in the following two epochs. Regarding the regression network training,  the maximum epoch is 50. It is worth mentioning that all the experimental pre-setting  will be fixed in all cross-dataset settings. 

Five evaluation metrics are reported for each experimental setting, including: Spearman rank correlation coefficient (SRCC), Pearson linear correlation coefﬁcient (PLCC), Kendall rank correlation coefficient (KRCC), Mean absolute error (MAE) and Root mean square error (RMSE). As suggested in~\cite{video2000final}, the predicted quality scores are passed through a nonlinear logistic mapping function before computing PLCC, RMSE and MAE:
\begin{equation}
\tilde{s}=\beta_{1}\left(\frac{1}{2}-\frac{1}{\exp \left(\beta_{2}\left(\hat{s}-\beta_{3}\right)\right)}\right)+\beta_{4} \hat{s}+\beta_{5}
\end{equation}
where \(\beta_{i}\) are regression parameters to be fitted.


\subsection{Quality Prediction Performance}
In this subsection, we evaluate the performance of our method with four different settings to further verify the effectiveness. We compare the proposed method with both conventional and deep learning based NR-IQA measures, including \bl{NIQE~\cite{mittal2012making}, PIQE~\cite{venkatanath2015blind}, SSEQ~\cite{liu2014no}, BRISQ~\cite{mittal2012no}, BPRI~\cite{min2017blind}, BMPRI~\cite{min2018blind}, HOSA~\cite{xu2016blind}, Rank~\cite{liu2017rankiqa}, MENO~\cite{ma2017end}, NRFR~\cite{bosse2017deep}, DB-CNN~\cite{zhang2018blind}.} Regarding the NRFR method, two modes are evaluated: patch-based and weighted patch based modes, which are further denoted as NRFR (patch) and NRFR (weighted), respectively. In particular, the conventional methods are pre-trained on NIs, and the deep learning based methods are trained with the data in the corresponding source domain. 

First, we treat TID2013 dataset as the source domain and SIQAD dataset as the target domain to train our model. The results are shown in Table~\ref{tab:siq}, from which we can find our method can achieve the best performance and the highest PLCC results demonstrate that the proposed method possesses a stronger linear relationship between quality prediction and MOS, comparing with all conventional methods and deep learning based methods.
To explore the influence of the source domain, we also conduct the experiments to replace the TID2013 with LIVE dataset. From Table~\ref{tab:siq}, we can find that our method can still acquire the best performance. However, compared with TID2013 as the source domain, the performance has been degraded to some extent as there are more distortion types involved in the TID2013 dataset. As such, more distortion relevant prior knowledge can be transferred to the target domain. {In Table~\ref{tab:dist}, we explore the performance of our method for each distortion type. As shown in the table, we find that the promising performance of our model can be achieved on the JPEG compression distortion and motion blur distortion. However, the performance is obviously degraded on the Gaussian noise. This is reasonable due to the rich textual information contained in SCIs. Compared with NIs, the injected Gaussian noise could cause more significant information loss, which could be much more sensitive to HVS in the perception of SCIs.}
\bl{For conventional NRIQA methods, we can observe that the HOSA and BMPRI models that are pretrained on the NI database have  a significant performance degradation on SCI databases,  validating the large domain gap between NIs and SCIs. Although the opinion-unaware model BPRI achieves the better performance on SCI databases, the proposed scheme still can outperform this method with a larger leap, demonstrating the effectiveness of our UDA strategy. }

To further explore the generalization ability of our method, we adopt another SCI dataset QACS as our target domain. As discussed in Section 4.1, this dataset considers two distortion types: HEVC and SCC extension based on HEVC, such that the distortions injected to these SCIs are closer to real application scenarios. The experimental results are shown in Tables~\ref{tab:hevc} and Tables~\ref{tab:scc}. Compared with the results of Table~\ref{tab:siq}, we can find our method still leads the performance by a large margin. Besides, we can find the performance improvement on the QACS dataset is much larger than on the SIQAD dataset, as only one distortion type exists in the target domain QACS resulting in the fact that feature centers can be  more easily learned.
Although the DB-CNN method achieves the second best results on QACS (HEVC) database, we argue that the method is a heavy-weight based method as two networks (SCNN and VGG16) are adopted. By contrast, our method only utilizes the SCNN which is much lighter than the DB-CNN method. 
\bl{Another phenomenon should also be noted is that our model can also achieve superior performances on the SCIs with distortion types unshared with  the source domain, revealing  a certain level of generalization capability of the proposed method.}

\begin{figure}[t]
\begin{minipage}[b]{0.85\linewidth}
  \centering
  \centerline{\includegraphics[width=1\linewidth]{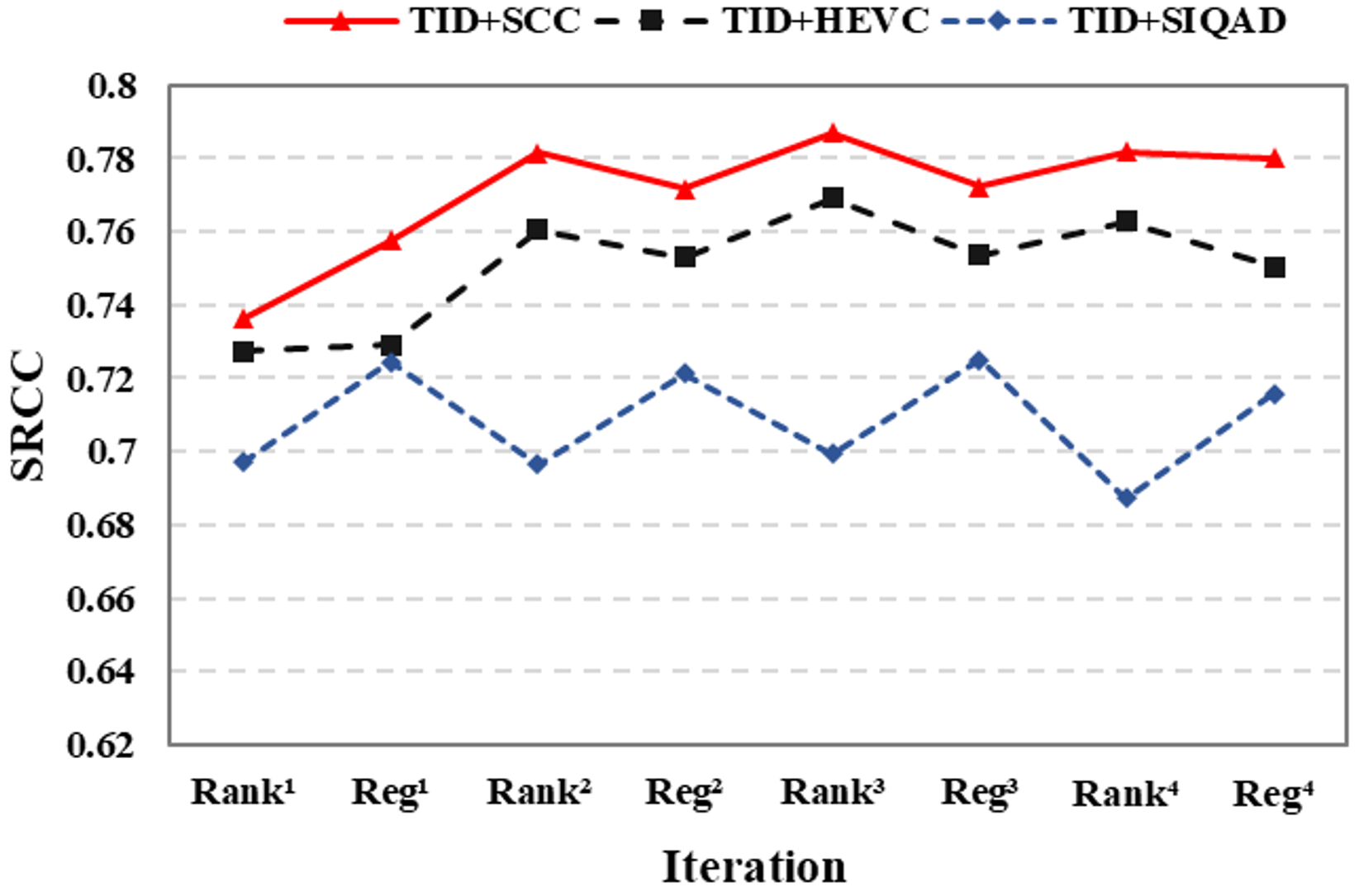}}
\end{minipage}
\caption{{The SRCC results variations with the number of iterations (source domain: TID2013, target domain: SIQAD). ``rank'' and ``reg'' indicate the results of the ranking model and regression model, respectively.}}
\label{tipplot}
\end{figure}

\begin{figure}[t]
\begin{minipage}[b]{0.85\linewidth}
  \centering
  \centerline{\includegraphics[width=1\linewidth]{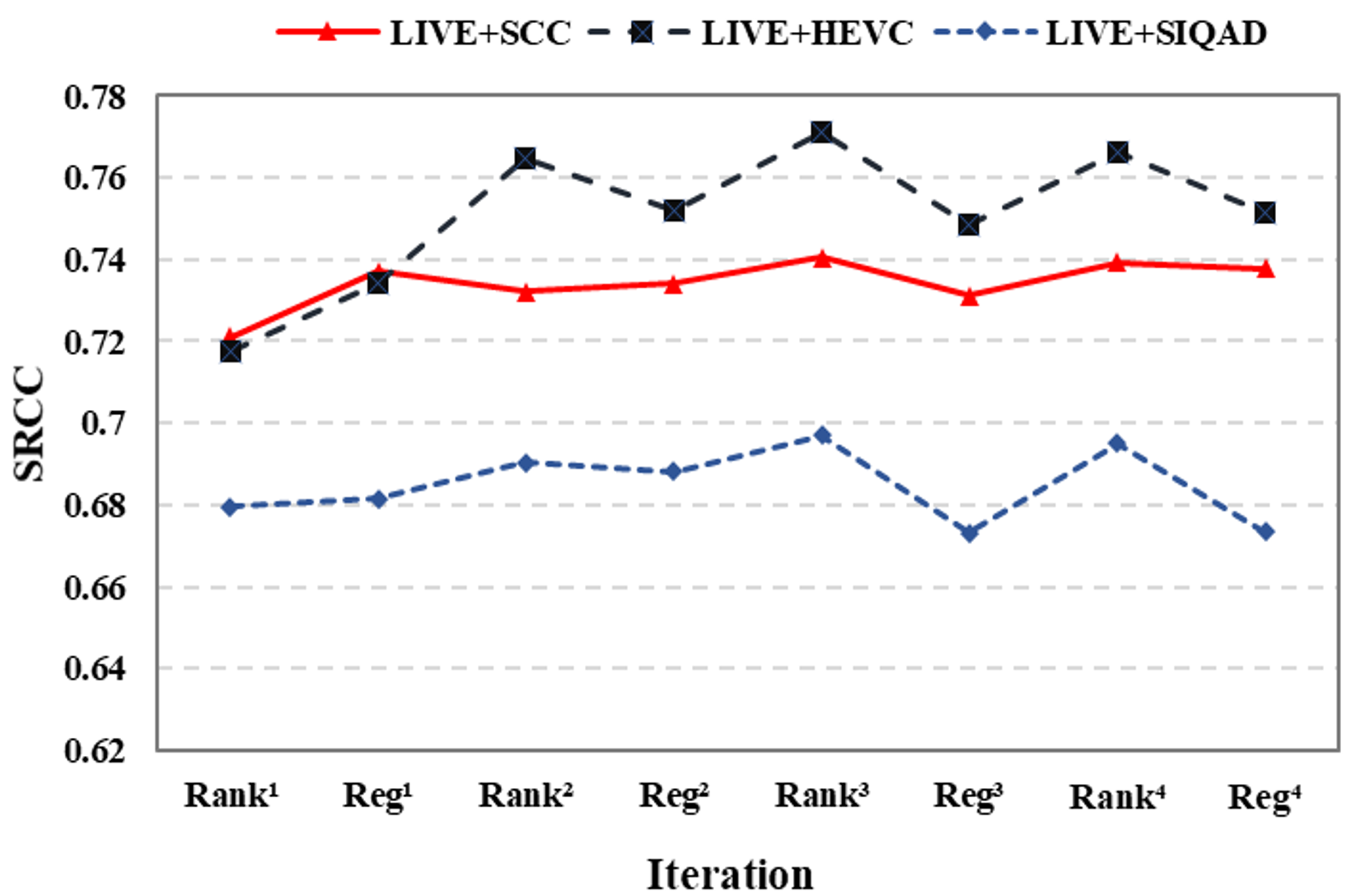}}
\end{minipage}
\caption{{The SRCC results variations with the number of iterations (source domain: LIVE, target domain: SIQAD). ``rank'' and ``reg'' indicate the results of the ranking model and regression model, respectively.}}
\label{liveplot}
\end{figure}

{In Figs.~\ref{tipplot} and~\ref{liveplot}, we investigate the influences of the number of iterations on the SRCC performance.  More specifically, TIP2013 and LIVE datasets are used as the source domains for training, respectively. From the two figures, we can find the SRCC of the ranking model in the second iteration tends to be saturated, although in some cases the performance can be further improved when the number of iterations has increased. Considering the trade-off between time-consumption and performance, we only train the ranking model twice and adopt the ranking results in the second iteration as our final result.}

\begin{table*}
  \centering
  \caption{Quality prediction performance comparisons (target domain: QACS HEVC).}
\begin{tabular}{c|c|ccccc}
\hline
\multicolumn{2}{c|}{Target domain: QACS (HEVC)} & PLCC  & MAE   & RMSE   & SRCC  & KRCC \bigstrut[t]\\
\hline
\multirow{4}[1]{*}{Conventional NIQA} & NIQE  & 0.3687  & 1.6259  & 1.9270  & 0.2724  & 0.1884  \\
      & PIQE  & 0.2084  & 1.7419  & 2.0276  & 0.2847  & 0.2071  \\
      & SSEQ  & 0.0712  & 1.7954  & 2.0678  & 0.0657  & 0.0453  \\
      & BRISQ & 0.4852  & 1.5144  & 1.8127  & 0.4788  & 0.3388  \\
      & HOSA     & -0.0082   & 1.8038    &2.0760    &0.0082    &0.0073  \\
& BMPRI  &0.2706    &1.7073    &1.9958   &0.1325   &0.0866  \\
& BPRI   &0.5146    &1.4635    &1.7775   &0.4941   &0.3408  \\
\hline
\multicolumn{1}{c|}{\multirow{6}[2]{*}{\shortstack{Deep Learning Based \\(Training on TID2013)}}} & Rank  & 0.3217  & 1.6538  & 1.9629  & 0.2577  & 0.1766  \bigstrut[t]\\
      & NRFR (weighted) & 0.1885  & 1.7608  & 2.0359  & 0.0853  & 0.0580  \\
      & NRFR (patch) & 0.1559  & 1.7713  & 2.0477  & 0.1347  & 0.0897  \\
      & MENO  & 0.3028  & 1.6732  & 1.9758  & 0.2453  & 0.1715  \\
      & DB-CNN  & 0.6321  & 1.2553  & 1.6064  & 0.6492  & 0.4735  \\
      & Ours  & \textbf{ 0.7658 } & \textbf{ 1.0052 } & \textbf{ 1.3332 } & \textbf{ 0.7605 } & \textbf{ 0.5844 } \bigstrut[b]\\
\hline
\multicolumn{1}{c|}{\multirow{5}[2]{*}{\shortstack{Deep Learning Based \\(Training on LIVE)}}} & Rank  & 0.4302  & 1.5722  & 1.8533  & 0.4011  & 0.3147  \bigstrut[t]\\
      & NRFR (weighted) & 0.2452  & 1.7576  & 2.0098  & 0.1938  & 0.1361  \\
      & NRFR (patch) & 0.2658  & 1.7173  & 1.9987  & 0.0570  & 0.0405  \\
      & DB-CNN  & 0.5103  & 1.3711  & 1.7320  & 0.5375  & 0.4417  \\
      & Ours  & \textbf{ 0.7995 } & \textbf{ 0.9769 } & \textbf{ 1.2452 } & \textbf{ 0.7646 } & \textbf{ 0.5898 } \bigstrut[b]\\
\hline
\end{tabular}%

  \label{tab:hevc}%

\end{table*}%

\begin{table*}
  \centering
  \caption{Quality prediction performance comparisons (target domain: QACS SCC).}
\begin{tabular}{c|c|ccccc}
\hline
\multicolumn{2}{c|}{Target domain: QACS (SCC)} & PLCC  & MAE   & RMSE   & SRCC  & KRCC \bigstrut[t]\\
\hline
\multirow{4}[1]{*}{Conventional NIQA} & NIQE  & 0.4973  & 1.6606  & 1.9652  & 0.4131  & 0.2903  \\
      & PIQE  & 0.0832  & 1.9139  & 2.2573  & 0.1717  & 0.1201  \\
      & SSEQ  & 0.0645  & 1.9172  & 2.2605  & 0.0669  & 0.0373  \\
      & BRISQ & 0.5058  & 1.6566  & 1.9541  & 0.5079  & 0.3628  \\
      & HOSA  &0.0727    &1.9230    &2.2592    &0.0553    &0.0390  \\
& BMPRI   &0.3310    &1.7934    &2.1375    &0.2638    &0.1860    \\
& BPRI   &0.6449    &1.4242    &1.7312   &0.5856   &0.4280\\
\hline
\multicolumn{1}{c|}{\multirow{6}[2]{*}{\shortstack{Deep Learning Based \\(Training on TID2013)}}} & Rank  & 0.4866  & 1.6579  & 1.9789  & 0.4761  & 0.3397  \bigstrut[t]\\
      & NRFR (weighted) & 0.3338  & 1.7839  & 2.1353  & 0.1051  & 0.0707  \\
      & NRFR (patch) & 0.2602  & 1.8159  & 2.1871  & 0.1668  & 0.1246  \\
      & MENO  & 0.1766  & 1.8945  & 2.2296  & 0.1392  & 0.0960  \\
      & DB-CNN & 0.6009  & 1.4327  & 1.8106  & 0.5954  & 0.4383  \\
      & Ours  & \textbf{ 0.7883 } & \textbf{ 1.0787 } & \textbf{ 1.3937 } & \textbf{ 0.7866 } & \textbf{ 0.5955 } \bigstrut[b]\\
\hline
\multicolumn{1}{c|}{\multirow{5}[2]{*}{\shortstack{Deep Learning Based \\(Training on LIVE)}}} & Rank  & 0.2377  & 1.7617  & 2.0001  & 0.1988  & 0.1462  \bigstrut[t]\\
      & NRFR (weighted) & 0.1753  & 1.8750  & 2.2301  & 0.0650  & 0.0453  \\
      & NRFR (patch) & 0.0723  & 1.9209  & 2.2592  & 0.0599  & 0.0405  \\
      & DB-CNN & 0.5247  & 1.6514  & 1.9371  & 0.5291  & 0.3800  \\
      & Ours  & \textbf{0.7500 } & \textbf{ 1.1498 } & \textbf{ 1.4983 } & \textbf{ 0.7322 } & \textbf{ 0.5557 } \bigstrut[b]\\
\hline
\end{tabular}%

  \label{tab:scc}%

\end{table*}%


\subsection{Ablation Study}
In this subsection, to reveal the functionalities of different modules in the proposed method, we perform the ablation study based on the cross-dataset setting (source domain: TID2013, target domain: QACS SCC), and the results are shown in Table~\ref{tab:ablation}.
\begin{table*}
  \centering
  \caption{Ablation studies on cross-dataset (training on TID2013, testing on QACS SCC).}
\begin{tabular}{c|ccccc}
\hline
Models & PLCC  & MAE   & RMSE   & SRCC  & KRCC \bigstrut\\
\hline
MOSPre. & 0.2012  & 1.8802  & 2.2189  & 0.0398  & 0.0264  \bigstrut[t]\\
MOSPre.+MMD & 0.4106  & 1.7301  & 2.0654  & 0.3789  & 0.2693  \\
RankPre. & 0.1779  & 1.8971  & 2.2291  & 0.0827  & 0.0601  \\
RankPre.+MMD & 0.2954  & 1.8395  & 2.1641  & 0.2414  & 0.1690  \\
RankPre.+MMD+Rec & 0.7374  & 1.1528  & 1.5123  & 0.7061  & 0.5312  \\
RankPre.+MMD+Rec+Cor (All) & 0.7672  & 1.1310  & 1.4529  & 0.7366  & 0.5604  \\
All+Retraining & 0.7883  & 1.0787  & 1.3937  & 0.7866  & 0.5955  \bigstrut[b]\\
\hline
\end{tabular}%

  \label{tab:ablation}%

\end{table*}%

\begin{table*}
  \centering
  \caption{Quality prediction performance of different cross-domain settings within CCT.}
\begin{tabular}{c|ccc|ccc}
\hline
\multicolumn{1}{c|}{\multirow{2}[4]{*}{Metric}} & \multicolumn{3}{c|}{With DA} & \multicolumn{3}{c}{Without DA} \\
\cline{2-7}      & \multicolumn{1}{l}{NI $\rightarrow$ SCI} & \multicolumn{1}{l}{SCI $\rightarrow$ CGI} & \multicolumn{1}{l|}{CGI $\rightarrow$ SCI} & \multicolumn{1}{l}{NI $\rightarrow$ SCI} & \multicolumn{1}{l}{SCI $\rightarrow$ CGI} & \multicolumn{1}{l}{CGI $\rightarrow$ SCI} \\
\hline
SRCC  & 0.7733 & 0.8209 & 0.7777 & 0.4372 & 0.6771 & 0.5669 \\
PLCC  & 0.7782 & 0.8244 & 0.7769 & 0.4670 & 0.6742 & 0.5894 \\
\hline
\end{tabular}%

  \label{tab:cct}%

\end{table*}%

{{In ablation studies, we first train the SCNN for quality prediction by the data in the source domain, leading to a model \textbf{MOSPre} which is directly applied on the target domain for testing. As expected, the performance is poor with high RMSE (2.2189) and extremely low SRCC (0.0398). This phenomenon reveals that the high domain gap exits between the two types of images and their relationship should be further exploited for model transfer.} Subsequently, we explore the DA based on the  \textbf{MOSPre} model by imposing the MMD loss, leading to the model \textbf{MOSPre+MMD}. We can find the performance is improved compared to \textbf{MOSPre} due to the fact that domains can be aligned when the MMD loss is introduced. However, the SRCC and PLCC are still not satisfactory, indicating that the MOS prediction model trained on the NIs is difficult to be straightforwardly transferred to SCIs due to the large discrepancy of deterministic or statistical characteristics relevant to quality. On the contrary, we adopt the ranking mechanism for quality transfer. Although the models (\textbf{RankPre} and \textbf{RankPre+MMD}) trained for pairwise ranking cannot achieve the same level of accuracy of \textbf{MOSPre+MMD} on QACS SCC, the transferring performance can be significantly improved when the classifier rectification loss is introduced in \textbf{MOSPre+MMD+Rec}. Moreover, based on \textbf{MOSPre+MMD+Rec}, we find our model can be further enhanced when we penalize the correlations of the learned ranking feature, which is denoted as \textbf{MOSPre+MMD+Rec+Cor (All)}, as more diversified quality relevant features with lower redundancy can be extracted. With the ranking model, the predicted quality score can be estimated and further used as the pseudo-labels to train the SCI specific regression network. As such, more reliable discriminative SCI pairs can be selected and the ranking-regression iteration can be performed. Finally, the best result can be achieved by our final model \textbf{All+Retraining}, as shown in Table~\ref{tab:ablation}.}

\subsection{Generalization Capability  Study on Different Content Types}
\bl{To verify the generalization capability of our UDA framework on different content types, we further performed the cross-content experiments on CCT database~\cite{min2017unified}. In particular, the  CCT database includes 3 content types (NI, SCI and CGI) and 2 distortion types (HEVC and SCC). To conduct the cross-content experiments, we select the images of all the three content types and the distortion type is fixed as HEVC. Then we evaluate our framework on three different cross-content settings, including: training on NIs and testing on SCIs (denoted as NI $\rightarrow$ SCI ),  training on SCIs and testing on CGIs (denoted as SCI $\rightarrow$ CGI) and  training on CGIs and testing on SCIs (denoted as CGI  $\rightarrow$ SCI). The results are shown in Table~\ref{tab:cct}. Moreover, we also provide the testing results without our UDA strategy. In other words, in this scenario,  we directly use the ranking model trained on source domain for evaluation on target domain. From the table, we can observe that significant performance improvement can be achieved by our proposed UDA strategy, revealing the knowledge of quality assessment model learned in one content type can be transferred to another content type, thereby improving the quality 
prediction accuracy on unlabeled data.}

\section{Conclusions}
We have presented a new NRIQA method based on unsupervised domain adaptation, in an effort to quest the transfer capability of the natural image quality. 
The proposed method is grounded on the unsupervised domain adaptation, equips the transferability of pair-wise relationship, and performs well on the target domain for specific application scenarios. The proposed method attempts to fill the gap between the statistics of SCIs and NIs through the ranking based relationship modeling, and the loss functions that minimize the feature discrepancy and rectify the classifier to accommodate the target domain lead to noticeable performance improvement in terms of the prediction accuracy. 

Recent years have witnessed a surge of images/videos that are not purely generated from optical cameras, such as
screen, gaming and mixture content. In particular, with the fast development of artificial intelligence, there are also numerous images and videos generated with the aid of deep generative network. As such, it is expected that the methodology and philosophy of the proposed method could play important roles in predicting the quality of these emerging domains. 
\bl{For example, rather than providing a DA based SCI quality measure only, we would also like to emphasize that the generalization capability of the NRIQA models could be further improved relying on investigation of the shareable knowledge and priors between different domains. Furthermore, the knowledge of quality assessment on visual images can be transferred to the other modalities.} It is also of interest to extend the current approach to other transferable tasks in quality assessment, such as distortion type and viewing condition. Moreover, it is imperative to study the quality assessment in the scenario that only a few samples are labelled with subjective ratings in the target domain, to meet the grand challenges faced by NRIQA in different real-world applications.

\bibliographystyle{IEEEtran}
\bibliography{IQACross}

\begin{thebibliography}{10}
\providecommand{\url}[1]{#1}
\csname url@samestyle\endcsname
\providecommand{\newblock}{\relax}
\providecommand{\bibinfo}[2]{#2}
\providecommand{\BIBentrySTDinterwordspacing}{\spaceskip=0pt\relax}
\providecommand{\BIBentryALTinterwordstretchfactor}{4}
\providecommand{\BIBentryALTinterwordspacing}{\spaceskip=\fontdimen2\font plus
\BIBentryALTinterwordstretchfactor\fontdimen3\font minus
  \fontdimen4\font\relax}
\providecommand{\BIBforeignlanguage}[2]{{%
\expandafter\ifx\csname l@#1\endcsname\relax
\typeout{** WARNING: IEEEtran.bst: No hyphenation pattern has been}%
\typeout{** loaded for the language `#1'. Using the pattern for}%
\typeout{** the default language instead.}%
\else
\language=\csname l@#1\endcsname
\fi
#2}}
\providecommand{\BIBdecl}{\relax}
\BIBdecl

\bibitem{moorthy2011blind}
A.~K. Moorthy and A.~C. Bovik, ``Blind image quality assessment: From natural
  scene statistics to perceptual quality,'' \emph{IEEE transactions on Image
  Processing}, vol.~20, no.~12, pp. 3350--3364, 2011.

\bibitem{gu2014using}
K.~Gu, G.~Zhai, X.~Yang, and W.~Zhang, ``Using free energy principle for blind
  image quality assessment,'' \emph{IEEE Transactions on Multimedia}, vol.~17,
  no.~1, pp. 50--63, 2014.

\bibitem{mittal2012no}
A.~Mittal, A.~K. Moorthy, and A.~C. Bovik, ``No-reference image quality
  assessment in the spatial domain,'' \emph{IEEE Transactions on image
  processing}, vol.~21, no.~12, pp. 4695--4708, 2012.

\bibitem{kang2014convolutional}
L.~Kang, P.~Ye, Y.~Li, and D.~Doermann, ``Convolutional neural networks for
  no-reference image quality assessment,'' in \emph{Proceedings of the IEEE
  conference on computer vision and pattern recognition}, 2014, pp. 1733--1740.

\bibitem{liu2018blind}
Y.~Liu, K.~Gu, S.~Wang, D.~Zhao, and W.~Gao, ``Blind quality assessment of
  camera images based on low-level and high-level statistical features,''
  \emph{IEEE Transactions on Multimedia}, vol.~21, no.~1, pp. 135--146, 2018.

\bibitem{wu2015blind}
Q.~Wu, H.~Li, F.~Meng, K.~N. Ngan, B.~Luo, C.~Huang, and B.~Zeng, ``Blind image
  quality assessment based on multichannel feature fusion and label transfer,''
  \emph{IEEE Transactions on Circuits and Systems for Video Technology},
  vol.~26, no.~3, pp. 425--440, 2015.

\bibitem{friston2006free}
K.~Friston, J.~Kilner, and L.~Harrison, ``A free energy principle for the
  brain,'' \emph{Journal of Physiology-Paris}, vol. 100, no. 1-3, pp. 70--87,
  2006.

\bibitem{friston2010free}
K.~Friston, ``The free-energy principle: a unified brain theory?'' \emph{Nature
  reviews neuroscience}, vol.~11, no.~2, pp. 127--138, 2010.

\bibitem{zhai2011psychovisual}
G.~Zhai, X.~Wu, X.~Yang, W.~Lin, and W.~Zhang, ``A psychovisual quality metric
  in free-energy principle,'' \emph{IEEE Transactions on Image Processing},
  vol.~21, no.~1, pp. 41--52, 2011.

\bibitem{zhai2019free}
G.~Zhai, X.~Min, and N.~Liu, ``Free-energy principle inspired visual quality
  assessment: An overview,'' \emph{Digital Signal Processing}, vol.~91, pp.
  11--20, 2019.

\bibitem{liu2019unsupervised}
Y.~Liu, K.~Gu, Y.~Zhang, X.~Li, G.~Zhai, D.~Zhao, and W.~Gao, ``Unsupervised
  blind image quality evaluation via statistical measurements of structure,
  naturalness, and perception,'' \emph{IEEE Transactions on Circuits and
  Systems for Video Technology}, vol.~30, no.~4, pp. 929--943, 2019.

\bibitem{min2020study}
X.~Min, G.~Zhai, J.~Zhou, M.~C. Farias, and A.~C. Bovik, ``Study of subjective
  and objective quality assessment of audio-visual signals,'' \emph{IEEE
  Transactions on Image Processing}, vol.~29, pp. 6054--6068, 2020.

\bibitem{kim2016fully}
J.~Kim and S.~Lee, ``Fully deep blind image quality predictor,'' \emph{IEEE
  Journal of selected topics in signal processing}, vol.~11, no.~1, pp.
  206--220, 2016.

\bibitem{ma2017end}
K.~Ma, W.~Liu, K.~Zhang, Z.~Duanmu, Z.~Wang, and W.~Zuo, ``End-to-end blind
  image quality assessment using deep neural networks,'' \emph{IEEE
  Transactions on Image Processing}, vol.~27, no.~3, pp. 1202--1213, 2017.

\bibitem{zhu2020multiple}
W.~Zhu, G.~Zhai, Z.~Han, X.~Min, T.~Wang, Z.~Zhang, and X.~Yangand, ``A
  multiple attributes image quality database for smartphone camera photo
  quality assessment,'' in \emph{2020 IEEE International Conference on Image
  Processing (ICIP)}.\hskip 1em plus 0.5em minus 0.4em\relax IEEE, 2020, pp.
  2990--2994.

\bibitem{zhang2021uncertainty}
W.~Zhang, K.~Ma, G.~Zhai, and X.~Yang, ``Uncertainty-aware blind image quality
  assessment in the laboratory and wild,'' \emph{IEEE Transactions on Image
  Processing}, vol.~30, pp. 3474--3486, 2021.

\bibitem{lu2020automatic}
Q.~Lu, G.~Zhai, W.~Zhu, Y.~Zhu, X.~Min, X.-P. Zhang, and H.~Yang, ``Automatic
  region selection for objective sharpness assessment of mobile device
  photos,'' in \emph{2020 IEEE International Conference on Image Processing
  (ICIP)}.\hskip 1em plus 0.5em minus 0.4em\relax IEEE, 2020, pp. 106--110.

\bibitem{wang2016just}
S.~Wang, L.~Ma, Y.~Fang, W.~Lin, S.~Ma, and W.~Gao, ``Just noticeable
  difference estimation for screen content images,'' \emph{IEEE Transactions on
  Image Processing}, vol.~25, no.~8, pp. 3838--3851, 2016.

\bibitem{min2017unified}
X.~Min, K.~Ma, K.~Gu, G.~Zhai, Z.~Wang, and W.~Lin, ``Unified blind quality
  assessment of compressed natural, graphic, and screen content images,''
  \emph{IEEE Transactions on Image Processing}, vol.~26, no.~11, pp.
  5462--5474, 2017.

\bibitem{ben2007analysis}
S.~Ben-David, J.~Blitzer, K.~Crammer, and F.~Pereira, ``Analysis of
  representations for domain adaptation,'' in \emph{Advances in neural
  information processing systems}, 2007, pp. 137--144.

\bibitem{ben2010theory}
S.~Ben-David, J.~Blitzer, K.~Crammer, A.~Kulesza, F.~Pereira, and J.~W.
  Vaughan, ``A theory of learning from different domains,'' \emph{Machine
  learning}, vol.~79, no. 1-2, pp. 151--175, 2010.

\bibitem{fernando2013unsupervised}
B.~Fernando, A.~Habrard, M.~Sebban, and T.~Tuytelaars, ``Unsupervised visual
  domain adaptation using subspace alignment,'' in \emph{Proceedings of the
  IEEE international conference on computer vision}, 2013, pp. 2960--2967.

\bibitem{saad2012blind}
M.~A. Saad, A.~C. Bovik, and C.~Charrier, ``Blind image quality assessment: A
  natural scene statistics approach in the dct domain,'' \emph{IEEE
  transactions on Image Processing}, vol.~21, no.~8, pp. 3339--3352, 2012.

\bibitem{hou2014blind}
W.~Hou, X.~Gao, D.~Tao, and X.~Li, ``Blind image quality assessment via deep
  learning,'' \emph{IEEE transactions on neural networks and learning systems},
  vol.~26, no.~6, pp. 1275--1286, 2014.

\bibitem{bosse2017deep}
S.~Bosse, D.~Maniry, K.-R. M{\"u}ller, T.~Wiegand, and W.~Samek, ``Deep neural
  networks for no-reference and full-reference image quality assessment,''
  \emph{IEEE Transactions on Image Processing}, vol.~27, no.~1, pp. 206--219,
  2017.

\bibitem{bianco2018use}
S.~Bianco, L.~Celona, P.~Napoletano, and R.~Schettini, ``On the use of deep
  learning for blind image quality assessment,'' \emph{Signal, Image and Video
  Processing}, vol.~12, no.~2, pp. 355--362, 2018.

\bibitem{gu2019blind}
J.~Gu, G.~Meng, S.~Xiang, and C.~Pan, ``Blind image quality assessment via
  learnable attention-based pooling,'' \emph{Pattern Recognition}, vol.~91, pp.
  332--344, 2019.

\bibitem{fu2016blind}
J.~Fu, H.~Wang, and L.~Zuo, ``Blind image quality assessment for multiply
  distorted images via convolutional neural networks,'' in \emph{2016 IEEE
  International Conference on Acoustics, Speech and Signal Processing
  (ICASSP)}.\hskip 1em plus 0.5em minus 0.4em\relax IEEE, 2016, pp. 1075--1079.

\bibitem{kim2018multiple}
J.~Kim, A.-D. Nguyen, S.~Ahn, C.~Luo, and S.~Lee, ``Multiple level
  feature-based universal blind image quality assessment model,'' in \emph{2018
  25th IEEE International Conference on Image Processing (ICIP)}.\hskip 1em
  plus 0.5em minus 0.4em\relax IEEE, 2018, pp. 291--295.

\bibitem{zhang2018blind}
W.~Zhang, K.~Ma, J.~Yan, D.~Deng, and Z.~Wang, ``Blind image quality assessment
  using a deep bilinear convolutional neural network,'' \emph{IEEE Transactions
  on Circuits and Systems for Video Technology}, 2018.

\bibitem{liu2017rankiqa}
X.~Liu, J.~van~de Weijer, and A.~D. Bagdanov, ``Rankiqa: Learning from rankings
  for no-reference image quality assessment,'' in \emph{Proceedings of the IEEE
  International Conference on Computer Vision}, 2017, pp. 1040--1049.

\bibitem{gu2017no}
K.~Gu, J.~Zhou, J.-F. Qiao, G.~Zhai, W.~Lin, and A.~C. Bovik, ``No-reference
  quality assessment of screen content pictures,'' \emph{IEEE Transactions on
  Image Processing}, vol.~26, no.~8, pp. 4005--4018, 2017.

\bibitem{fang2017no}
Y.~Fang, J.~Yan, L.~Li, J.~Wu, and W.~Lin, ``No reference quality assessment
  for screen content images with both local and global feature
  representation,'' \emph{IEEE Transactions on Image Processing}, vol.~27,
  no.~4, pp. 1600--1610, 2017.

\bibitem{zheng2019no}
L.~Zheng, L.~Shen, J.~Chen, P.~An, and J.~Luo, ``No-reference quality
  assessment for screen content images based on hybrid region features
  fusion,'' \emph{IEEE Transactions on Multimedia}, vol.~21, no.~8, pp.
  2057--2070, 2019.

\bibitem{zuo2016screen}
L.~Zuo, H.~Wang, and J.~Fu, ``Screen content image quality assessment via
  convolutional neural network,'' in \emph{2016 IEEE International Conference
  on Image Processing (ICIP)}.\hskip 1em plus 0.5em minus 0.4em\relax IEEE,
  2016, pp. 2082--2086.

\bibitem{min2018saliency}
X.~Min, K.~Gu, G.~Zhai, M.~Hu, and X.~Yang, ``Saliency-induced
  reduced-reference quality index for natural scene and screen content
  images,'' \emph{Signal Processing}, vol. 145, pp. 127--136, 2018.

\bibitem{min2017blind}
X.~Min, K.~Gu, G.~Zhai, J.~Liu, X.~Yang, and C.~W. Chen, ``Blind quality
  assessment based on pseudo-reference image,'' \emph{IEEE Transactions on
  Multimedia}, vol.~20, no.~8, pp. 2049--2062, 2017.

\bibitem{min2018blind}
X.~Min, G.~Zhai, K.~Gu, Y.~Liu, and X.~Yang, ``Blind image quality estimation
  via distortion aggravation,'' \emph{IEEE Transactions on Broadcasting},
  vol.~64, no.~2, pp. 508--517, 2018.

\bibitem{borgwardt2006integrating}
K.~M. Borgwardt, A.~Gretton, M.~J. Rasch, H.-P. Kriegel, B.~Sch{\"o}lkopf, and
  A.~J. Smola, ``Integrating structured biological data by kernel maximum mean
  discrepancy,'' \emph{Bioinformatics}, vol.~22, no.~14, pp. e49--e57, 2006.

\bibitem{huang2007correcting}
J.~Huang, A.~Gretton, K.~Borgwardt, B.~Sch{\"o}lkopf, and A.~J. Smola,
  ``Correcting sample selection bias by unlabeled data,'' in \emph{Advances in
  neural information processing systems}, 2007, pp. 601--608.

\bibitem{gong2013connecting}
B.~Gong, K.~Grauman, and F.~Sha, ``Connecting the dots with landmarks:
  Discriminatively learning domain-invariant features for unsupervised domain
  adaptation,'' in \emph{International Conference on Machine Learning}, 2013,
  pp. 222--230.

\bibitem{long2017deep}
M.~Long, H.~Zhu, J.~Wang, and M.~I. Jordan, ``Deep transfer learning with joint
  adaptation networks,'' in \emph{Proceedings of the 34th International
  Conference on Machine Learning-Volume 70}.\hskip 1em plus 0.5em minus
  0.4em\relax JMLR. org, 2017, pp. 2208--2217.

\bibitem{yan2017mind}
H.~Yan, Y.~Ding, P.~Li, Q.~Wang, Y.~Xu, and W.~Zuo, ``Mind the class weight
  bias: Weighted maximum mean discrepancy for unsupervised domain adaptation,''
  in \emph{Proceedings of the IEEE Conference on Computer Vision and Pattern
  Recognition}, 2017, pp. 2272--2281.

\bibitem{sun2016deep}
B.~Sun and K.~Saenko, ``Deep coral: Correlation alignment for deep domain
  adaptation,'' in \emph{European conference on computer vision}.\hskip 1em
  plus 0.5em minus 0.4em\relax Springer, 2016, pp. 443--450.

\bibitem{peng2018synthetic}
X.~Peng and K.~Saenko, ``Synthetic to real adaptation with generative
  correlation alignment networks,'' in \emph{2018 IEEE Winter Conference on
  Applications of Computer Vision (WACV)}.\hskip 1em plus 0.5em minus
  0.4em\relax IEEE, 2018, pp. 1982--1991.

\bibitem{zhuang2015supervised}
F.~Zhuang, X.~Cheng, P.~Luo, S.~J. Pan, and Q.~He, ``Supervised representation
  learning: Transfer learning with deep autoencoders,'' in \emph{Twenty-Fourth
  International Joint Conference on Artificial Intelligence}, 2015.

\bibitem{li2020unsupervised}
H.~Li, R.~Wan, S.~Wang, and A.~C. Kot, ``Unsupervised domain adaptation in the
  wild via disentangling representation learning,'' \emph{International Journal
  of Computer Vision}, pp. 1--17, 2020.

\bibitem{das2018sample}
D.~Das and C.~G. Lee, ``Sample-to-sample correspondence for unsupervised domain
  adaptation,'' \emph{Engineering Applications of Artificial Intelligence},
  vol.~73, pp. 80--91, 2018.

\bibitem{das2018graph}
------, ``Graph matching and pseudo-label guided deep unsupervised domain
  adaptation,'' in \emph{International conference on artificial neural
  networks}.\hskip 1em plus 0.5em minus 0.4em\relax Springer, 2018, pp.
  342--352.

\bibitem{das2018unsupervised}
------, ``Unsupervised domain adaptation using regularized hyper-graph
  matching,'' in \emph{2018 25th IEEE International Conference on Image
  Processing (ICIP)}.\hskip 1em plus 0.5em minus 0.4em\relax IEEE, 2018, pp.
  3758--3762.

\bibitem{ganin2014unsupervised}
Y.~Ganin and V.~Lempitsky, ``Unsupervised domain adaptation by
  backpropagation,'' \emph{arXiv preprint arXiv:1409.7495}, 2014.

\bibitem{taigman2016unsupervised}
Y.~Taigman, A.~Polyak, and L.~Wolf, ``Unsupervised cross-domain image
  generation,'' \emph{arXiv preprint arXiv:1611.02200}, 2016.

\bibitem{bousmalis2017unsupervised}
K.~Bousmalis, N.~Silberman, D.~Dohan, D.~Erhan, and D.~Krishnan, ``Unsupervised
  pixel-level domain adaptation with generative adversarial networks,'' in
  \emph{Proceedings of the IEEE conference on computer vision and pattern
  recognition}, 2017, pp. 3722--3731.

\bibitem{mittal2012making}
A.~Mittal, R.~Soundararajan, and A.~C. Bovik, ``Making a “completely blind”
  image quality analyzer,'' \emph{IEEE Signal Processing Letters}, vol.~20,
  no.~3, pp. 209--212, 2012.

\bibitem{ponomarenko2015image}
N.~Ponomarenko, L.~Jin, O.~Ieremeiev, V.~Lukin, K.~Egiazarian, J.~Astola,
  B.~Vozel, K.~Chehdi, M.~Carli, F.~Battisti \emph{et~al.}, ``Image database
  tid2013: Peculiarities, results and perspectives,'' \emph{Signal Processing:
  Image Communication}, vol.~30, pp. 57--77, 2015.

\bibitem{yang2015perceptual}
H.~Yang, Y.~Fang, and W.~Lin, ``Perceptual quality assessment of screen content
  images,'' \emph{IEEE Transactions on Image Processing}, vol.~24, no.~11, pp.
  4408--4421, 2015.

\bibitem{bengio2013representation}
Y.~Bengio, A.~Courville, and P.~Vincent, ``Representation learning: A review
  and new perspectives,'' \emph{IEEE transactions on pattern analysis and
  machine intelligence}, vol.~35, no.~8, pp. 1798--1828, 2013.

\bibitem{simonyan2014very}
K.~Simonyan and A.~Zisserman, ``Very deep convolutional networks for
  large-scale image recognition,'' \emph{arXiv preprint arXiv:1409.1556}, 2014.

\bibitem{he2016deep}
K.~He, X.~Zhang, S.~Ren, and J.~Sun, ``Deep residual learning for image
  recognition,'' in \emph{Proceedings of the IEEE conference on computer vision
  and pattern recognition}, 2016, pp. 770--778.

\bibitem{breiman1996bagging}
L.~Breiman, ``Bagging predictors,'' \emph{Machine learning}, vol.~24, no.~2,
  pp. 123--140, 1996.

\bibitem{cogswell2015reducing}
M.~Cogswell, F.~Ahmed, R.~Girshick, L.~Zitnick, and D.~Batra, ``Reducing
  overfitting in deep networks by decorrelating representations,'' \emph{arXiv
  preprint arXiv:1511.06068}, 2015.

\bibitem{rodriguez2016regularizing}
P.~Rodr{\'\i}guez, J.~Gonzalez, G.~Cucurull, J.~M. Gonfaus, and X.~Roca,
  ``Regularizing cnns with locally constrained decorrelations,'' \emph{arXiv
  preprint arXiv:1611.01967}, 2016.

\bibitem{gretton2012kernel}
A.~Gretton, K.~M. Borgwardt, M.~J. Rasch, B.~Sch{\"o}lkopf, and A.~Smola, ``A
  kernel two-sample test,'' \emph{Journal of Machine Learning Research},
  vol.~13, no. Mar, pp. 723--773, 2012.

\bibitem{wen2016discriminative}
Y.~Wen, K.~Zhang, Z.~Li, and Y.~Qiao, ``A discriminative feature learning
  approach for deep face recognition,'' in \emph{European conference on
  computer vision}.\hskip 1em plus 0.5em minus 0.4em\relax Springer, 2016, pp.
  499--515.

\bibitem{ma2017dipiq}
K.~Ma, W.~Liu, T.~Liu, Z.~Wang, and D.~Tao, ``dipiq: Blind image quality
  assessment by learning-to-rank discriminable image pairs,'' \emph{IEEE
  Transactions on Image Processing}, vol.~26, no.~8, pp. 3951--3964, 2017.

\bibitem{sheikh2003image}
H.~R. Sheikh, ``Image and video quality assessment research at live,''
  \emph{http://live. ece. utexas. edu/research/quality}, 2003.

\bibitem{wang2016subjective}
S.~Wang, K.~Gu, X.~Zhang, W.~Lin, L.~Zhang, S.~Ma, and W.~Gao, ``Subjective and
  objective quality assessment of compressed screen content images,''
  \emph{IEEE Journal on Emerging and Selected Topics in Circuits and Systems},
  vol.~6, no.~4, pp. 532--543, 2016.

\bibitem{sullivan2012overview}
G.~J. Sullivan, J.-R. Ohm, W.-J. Han, and T.~Wiegand, ``Overview of the high
  efficiency video coding (hevc) standard,'' \emph{IEEE Transactions on
  circuits and systems for video technology}, vol.~22, no.~12, pp. 1649--1668,
  2012.

\bibitem{shi2015study}
S.~Shi, X.~Zhang, S.~Wang, R.~Xiong, and S.~Ma, ``Study on subjective quality
  assessment of screen content images,'' in \emph{2015 Picture Coding Symposium
  (PCS)}.\hskip 1em plus 0.5em minus 0.4em\relax IEEE, 2015, pp. 75--79.

\bibitem{2004Image}
Z.~Wang, ``Image quality assessment : From error visibility to structural
  similarity,'' \emph{IEEE Transactions on Image Processing}, 2004.

\bibitem{paszke2019pytorch}
A.~Paszke, S.~Gross, F.~Massa, A.~Lerer, J.~Bradbury, G.~Chanan, T.~Killeen,
  Z.~Lin, N.~Gimelshein, L.~Antiga \emph{et~al.}, ``Pytorch: An imperative
  style, high-performance deep learning library,'' in \emph{Advances in neural
  information processing systems}, 2019, pp. 8026--8037.

\bibitem{prechelt1998early}
L.~Prechelt, ``Early stopping-but when?'' in \emph{Neural Networks: Tricks of
  the trade}.\hskip 1em plus 0.5em minus 0.4em\relax Springer, 1998, pp.
  55--69.

\bibitem{video2000final}
V.~Q.~E. Group \emph{et~al.}, ``Final report from the video quality experts
  group on the validation of objective models of video quality assessment,'' in
  \emph{VQEG meeting, Ottawa, Canada, March, 2000}, 2000.

\bibitem{venkatanath2015blind}
N.~Venkatanath, D.~Praneeth, M.~C. Bh, S.~S. Channappayya, and S.~S. Medasani,
  ``Blind image quality evaluation using perception based features,'' in
  \emph{2015 Twenty First National Conference on Communications (NCC)}.\hskip
  1em plus 0.5em minus 0.4em\relax IEEE, 2015, pp. 1--6.

\bibitem{liu2014no}
L.~Liu, B.~Liu, H.~Huang, and A.~C. Bovik, ``No-reference image quality
  assessment based on spatial and spectral entropies,'' \emph{Signal
  Processing: Image Communication}, vol.~29, no.~8, pp. 856--863, 2014.

\bibitem{xu2016blind}
J.~Xu, P.~Ye, Q.~Li, H.~Du, Y.~Liu, and D.~Doermann, ``Blind image quality
  assessment based on high order statistics aggregation,'' \emph{IEEE
  Transactions on Image Processing}, vol.~25, no.~9, pp. 4444--4457, 2016.

\end{thebibliography}


\end{document}